\documentclass[lettersize,journal]{IEEEtran}
\usepackage[T1]{fontenc}
\usepackage[misc]{ifsym}
\usepackage{graphicx}
\usepackage{amssymb}
\usepackage{makecell}
\usepackage{algorithm}
\usepackage{algorithmic}
\usepackage{amsthm}
\newtheorem{definition}{Definition}
\usepackage{amsmath,amsfonts}
\usepackage{multirow}
\usepackage{array}
\usepackage{diagbox}
\usepackage[symbol]{footmisc}
\usepackage{subfigure}

\begin{document}

\title{Extracting Spatiotemporal Data from Gradients with Large Language Models}

\author{Lele Zheng, Yang Cao, Renhe Jiang, Kenjiro Taura, Yulong Shen, Sheng Li, and Masatoshi Yoshikawa

\thanks{This work is supported in part by JST CREST (No. JPMJCR21M2); in part by JSPS KAKENHI (No. JP22H00521, JP22H03595, JP21K19767); in part by JST/NSF Joint Research SICORP (No. JPMJSC2107); in part by the National Natural Science Foundation of China (No. 62220106004, 61972308). \textit{(Corresponding author: Yang Cao.)}}
\thanks {Lele Zheng and Yang Cao are with the Department of Computer Science, Tokyo Institute of Technology, Tokyo, Japan (e-mail: llzhengstu@gmail.com; cao@c.titech.ac.jp).}
\thanks {Renhe Jiang is with the Center for Spatial Information Science, The University of Tokyo, Tokyo, Japan (e-mail: jiangrh@csis.u-tokyo.ac.jp).}
\thanks {Kenjiro Taura is with the Department of Information and Communication Engineering, The University of Tokyo, Tokyo, Japan (e-mail: tau@eidos.ic.i.u-tokyo.ac.jp).}
\thanks {Lele Zheng and Yulong Shen are with the School of Computer Science and Technology, Xidian University, Xi’an, Shaanxi, China (e-mail: ylshen@mail.xidian.edu.cn).}
\thanks {Sheng Li is with the National Institute of Information and Communications Technology, Kyoto, Japan (e-mail: sheng.li@nict.go.jp).}
\thanks {Masatoshi Yoshikawa is with the School Osaka
Seikei University, Osaka, Japan (e-mail: yoshikawa-mas@osaka-seikei.ac.jp).}
}

\maketitle

\begin{abstract}
Recent works show that sensitive user data can be reconstructed from gradient updates, breaking the key privacy promise of federated learning. While success was demonstrated primarily on image data, these methods do not directly transfer to other domains, such as spatiotemporal data. To understand privacy risks in spatiotemporal federated learning, we first propose Spatiotemporal Gradient Inversion Attack (ST-GIA), a gradient attack algorithm tailored to spatiotemporal data that successfully reconstructs the original location from gradients. Furthermore, the absence of priors in attacks on spatiotemporal data has hindered the accurate reconstruction of real client data. To address this limitation, we propose ST-GIA+, which utilizes an auxiliary language model to guide the search for potential locations, thereby successfully reconstructing the original data from gradients. In addition, we design an adaptive defense strategy to mitigate gradient inversion attacks in spatiotemporal federated learning. By dynamically adjusting the perturbation levels, we can offer tailored protection for varying rounds of training data, thereby achieving a better trade-off between privacy and utility than current state-of-the-art methods. Through intensive experimental analysis on three real-world datasets, we reveal that the proposed defense strategy can well preserve the utility of spatiotemporal federated learning with effective security protection.
\end{abstract}

\begin{IEEEkeywords}
Gradient inversion attacks, federated learning, spatiotemporal data, differential privacy, large language model
\end{IEEEkeywords}

\section{Introduction}\label{sec:intro}
Spatiotemporal data analysis tasks, such as human mobility predictions (HMP), are crucial in various fields due to their ability to predict and analyze movement patterns of individuals or groups~\cite{zhang2023federated}. Accurate human mobility predictions can improve urban planning~\cite{yuan2012discovering}, recommend relevant points of interest~\cite{liu2017point}, and enhance transportation systems in smart cities~\cite{shi2019survey}. To address privacy concerns, federated learning (FL) has been extensively employed in human mobility prediction, allowing clients to collaboratively train shared models without directly disclosing their private data~\cite{feng2020pmf,li2020predicting}.

\begin{figure*}
        \centering
    \includegraphics[width=0.9\textwidth]{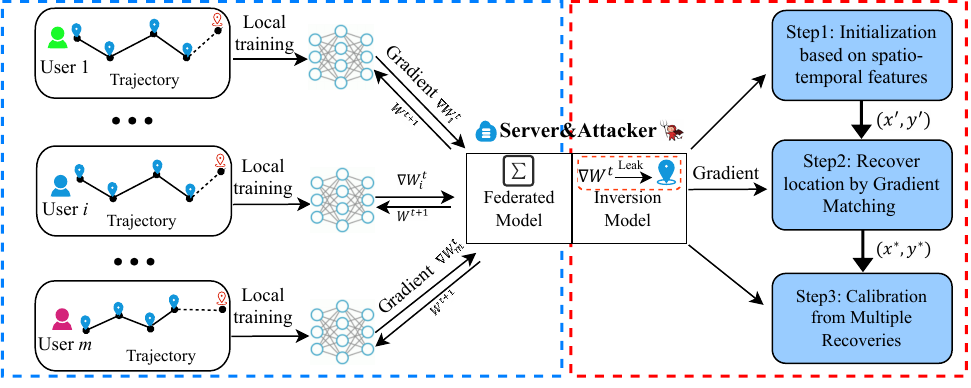}
    \caption{Overview of ST-GIA. The left part (blue box) performs the federated protocol, and the right part (red box) illustrates the main steps of ST-GIA.}
    \label{fig: system}
\end{figure*}

Although federated learning inherently enhances privacy by enabling clients to keep private data on their local devices, recent studies have highlighted the potential vulnerability of shared gradients to gradient inversion attacks~\cite{fan2024guardian,geiping2020inverting,geng2023improved,zhu2019deep}. As shown in Fig.~\ref{fig: system}, while ostensibly adhering to protocol, the honest-but-curious server might covertly steal privacy by leveraging gradient inversion attacks on gradients shared by clients, thereby reconstructing raw data. Most gradient inversion attacks primarily focus on minimizing the distance between the dummy gradients and the ground-truth gradients. To generate dummy gradients, random data and corresponding labels are fed to the global model. Taking the distance between the gradients as error and the dummy inputs as parameters, the recovery process can be formulated as an iterative optimization problem. Upon the convergence of this optimization process, the private data is expected to be comprehensively reconstructed. 

However, although the existing gradient inversion attacks have undeniable contributions, they still have some limitations that cannot be ignored. (1) First, the existing attacks are developed to reconstruct training \textit{images} or \textit{texts} used to train classifiers and have yet to be validated in spatiotemporal federated learning. (2) Second, we find that the use of prior knowledge can substantially enhance the attacker's capabilities in both the initialization and iterative optimization phases. Specifically, in the initialization phase, current gradient inversion attacks tend to analyze each round in isolation, overlooking the integration of vital information from the overall training process. Within the context of spatiotemporal federated learning, the outcome of an attack in a given round is likely to influence the attack result in the subsequent round significantly. In the optimization phase, further calibration of the optimization results based on prior knowledge can improve attack accuracy. (3) Third, in the context of spatiotemporal federated learning, achieving effective outcomes through the direct application of existing differential privacy-based defense methods proves challenging. The reason is that these methods are primarily designed for general-purpose defenses and are not tailored to address gradient leakage attacks specifically.

In this paper, we first propose a novel gradient attack algorithm, named ST-GIA, designed explicitly to raise awareness of spatiotemporal data, which can effectively reconstruct the original location from shared gradients. In ST-GIA, the attackers first exploit the characteristics of spatiotemporal data to initialize dummy data. Then, the attackers recover the original location through gradient matching, effectively leveraging a priori knowledge of the road network to significantly improve their attacks' accuracy. However, we find that using only spatiotemporal features and prior knowledge of the road network achieves better attack results when the number of global rounds is low. Conversely, attack results remain poor with a high number of global rounds because the gradient contains progressively less information as the global model converges. Therefore, we propose ST-GIA+, which utilizes an auxiliary language model to guide the search for potential locations, thereby successfully reconstructing the original data from the gradient. Finally, they employ multiple recovery results to refine and calibrate the final reconstruction outcomes. In addition, we evaluate the effectiveness of existing differential privacy-based defense methods against ST-GIA+ and propose a new adaptive privacy-preserving strategy tailored to mitigate gradient inversion attacks in spatiotemporal federated learning. In particular, we design an importance-aware budget allocation method to ensure sensitivity to different training rounds. The main contributions are summarized as follows:
\begin{enumerate}
    \item As the first attempt in the field of spatiotemporal federated learning, we propose ST-GIA, a gradient attack algorithm for spatiotemporal data that can effectively reconstruct the original location from shared gradients.
    \item We further propose ST-GIA+, which is an enhanced version of the spatiotemporal gradient inversion attack that utilizes an auxiliary language model to guide the search for potential locations, thereby successfully reconstructing the original data from gradients. 
    \item We propose a novel metric for evaluating the attack risk faced by spatiotemporal data. Based on this new metric, we enhance the privacy budget allocation method and design an adaptive privacy-preserving strategy tailored to mitigate gradient inversion attacks in spatiotemporal federated learning.
    \item Comprehensive experiments conducted on three real-world datasets demonstrate the effectiveness of the proposed attack and defense strategies. 
\end{enumerate}

\section{Related Work}
\subsection{Spatiotemporal Federated Learning}
Human mobility prediction is one of the most popular tasks in spatiotemporal federated learning and has been extensively studied in recent years~\cite{fan2019decentralized,feng2020pmf,li2020predicting,wang2022location,zhang2023federated}. For example, Li et al.~\cite{li2020predicting} developed a spatial-temporal self-attention network to integrate spatiotemporal information for enhanced location prediction. Feng et al.~\cite{feng2020pmf} introduced a privacy-preserving mobility prediction framework PMF using federated learning, which exhibited notable performance compared to centralized models. Fan et al.~\cite{fan2019decentralized} designed a decentralized attention-based personalized human mobility prediction model and implemented pre-training strategies to expedite the federated learning process. Li et al.~\cite{li2020predicting} proposed a spatiotemporal self-attention network (STSAN) for integrating spatiotemporal information for location prediction and designed an adaptive model fusion method to update the parameters under the federated learning framework. However, malicious attackers may steal private information from shared gradients through gradient inversion attacks.

\subsection{Gradient Inversion Attacks} 
Zhu et al.~\cite{zhu2019deep} demonstrated in their pioneering work that sharing gradients can still lead to data privacy leakage. They introduced an attack method called DLG, which can reconstruct original images and labels by minimizing the Euclidean distance between the true gradient and the dummy gradient generated from random noise data. They also showed that this attack worked well in the text domain. However, their method performs best with small images and small batch sizes. Building on this, Zhao et al.~\cite{zhao2020idlg} proposed iDLG, which can extract true labels from gradients, giving it an advantage over DLG, but it was only effective when the batch size equals one.
SAPAG~\cite{wang2020sapag}  introduced a more general attack method, using a Gaussian kernel based on gradient differences as a distance metric to speed up convergence. Geiping et al.~\cite{geiping2020inverting} improved image reconstruction quality by using cosine distance instead of Euclidean distance for gradient differences and applied total variation as a regularizer for image denoising. They also compared the impact of trained versus untrained models on image recovery. CPL~\cite{wei2020framework}  further enhanced attack effectiveness by using $L_2$ distance and label-based regularization.
Recently, Yin et al.~\cite{geng2023improved} extended the label extraction method to batch mode with GradInversion~\cite{yin2021see}. Their method assumed no label repetition within a mini-batch, but this assumption didn't hold in practice when batch sizes are large, and the number of classes is limited. They also demonstrated the impact of random seeds on recovery results and proposed a group consistency regularization method using the RANSAC flow algorithm. However, current works focus on reconstructing training images or texts and have yet to be validated in spatiotemporal federated learning.

\subsection{Large Language Models for Human Mobility Prediction}
Large Language Models (LLMs) have shown significant effectiveness in handling time-series data by leveraging their ability to capture complex patterns, dependencies, and temporal relationships. Xue et al.~\cite{xue2022translating} tackled human mobility forecasting by framing it as a natural language translation task, where mobility data is converted into sentences using a description template, and a sequence-to-sequence model predicts future mobility. The authors also proposed a two-branch SHIFT architecture to manage the translation process. Building on this, AuxMobLCast~\cite{xue2022leveraging} explored prompt engineering for time-series data. Chang et al.~\cite{chang2023llm4ts} introduced LLM4TS, a framework for time-series forecasting using pre-trained language models, which employs a two-stage fine-tuning process: first aligning the model with time-series data, followed by fine-tuning for forecasting. The framework also introduces a two-level aggregation method to better integrate and interpret multi-scale temporal data. Li et al.~\cite{li2024large} proposed a framework using large language models with commonsense knowledge for next point-of-interest recommendations, developing trajectory prompts to frame the task as question-answering and implementing key-query similarity to address the cold start problem. Similarly, Wang et al.~\cite{wang2023would} introduced LLM-Mob, a method utilizing large language models to analyze human mobility data. By incorporating historical and contextual stays, it captures both long and short-term movement patterns and enables time-aware predictions, with the addition of context-based prompts to improve prediction accuracy.

\section{Preliminaries}
\subsection{Spatiotemporal Federated Learning}
One typical task requiring spatiotemporal federated learning is human mobility prediction, which utilizes historical trajectories to predict the location $x_u^{n+1}$ of the target user (client) $u$ in the next time step, where spatiotemporal point $x_u$ can be described as a 3-tuple of the time stamp, latitude, and longitude. Thus, the raw data collected by user $u$ at time $t$ can be formally represented as follows:
    \begin{equation}
        x_u^t = {(t, lat, lon)}
    \end{equation}
The historical trajectory $s_u$ of each user $u$can be defined as:
    \begin{equation}
        s_u = \{x_u^{0}, x_u^{1}, \dots, x_u^{n} \}
    \end{equation}
where $x_u^{i}$ is the $i$-th record of user $u$, ordered by time, and $n$ is the number of trajectory points. In mobility trajectories, location identifiers can be latitude-longitude coordinates, spatial grid IDs, or street numbers. To simplify matters, recent studies have converted various location identifiers into a single unique identifier. Additionally, time intervals are quantized into fixed values to further streamline the modeling process. 

In such a scenario, spatiotemporal federated learning aggregates model parameters from different clients into a global model, where the clients learn the temporal and spatial correlations of the data locally. The client first downloads the global state $w_t$ from the server and then performs local training using the local trajectory data, i.e., $w_{t+1}=w_t-\eta\nabla w_t$, where $w_t$ is the local model parameter update at round $t$ and $\nabla w_t$ is the gradient of the model parameters. Upon receiving the local updates from clients, the server aggregates these updates to get a global state and initiates the next round of federated learning. 
Several models can be used for local training to reveal spatiotemporal correlations in trajectory data. For example, Long short-term memory (LSTM)~\cite{hochreiter1997long} is a special kind of RNN, which is designed to solve the gradient disappearance and gradient explosion problems during the training of long sequences. DeepMove~\cite{feng2018deepmove} is an attentional mobility model that includes a multi-modal embedding recurrent neural network for capturing multiple factors about human mobility and a historical attention module for modeling the multi-level periodicity of human mobility. STRNN~\cite{liu2016predicting} uses temporal and spatial intervals between every two consecutive visits as explicit information to improve model performance.

\subsection{Gradient Inversion Attack}
A gradient inversion attack is carried out by the server, or an entity that has compromised it, with the intention of acquiring a client's private data $(x^*, y^*)$ through the analysis of gradient updates 
$\nabla_{\theta^k}\mathcal{L}(x^*, y^*)$ uploaded to the server. These attacks typically presume the presence of honest-but-curious servers, which adhere to the federated training protocol without altering it. The common method for obtaining private data is to solve an optimization problem:
\begin{equation}
    \arg\min\limits_{(x', y')} \delta(\nabla_{\theta^k}\mathcal{L}(x^*, y^*), \nabla_{\theta^k}\mathcal{L}(x', y')),
\end{equation}
where $\delta$ represents a specific distance measure and $(x', y')$ denotes dummy data. Typical choices for $\delta$ are $L_2, L_1$, and cosine distances. 

\subsection{Local Differential Privacy}
Local differential privacy (LDP)~\cite{bassily2015local, wang2017locally} has emerged as the gold standard for protecting individual privacy in scenarios where user data is collected by an untrusted data collector. Essentially, LDP enables users to determine the extent to which their data is distinguishable to the data collector through a privacy parameter, $\epsilon$, chosen by the user.
\begin{definition}[Local Differential Privacy]
A randomized algorithm $\mathcal{M}$ satisfies $\epsilon$-local differential privacy if for any two inputs $x, x'\in \mathcal{D}$ and for any output $y\in \mathcal{Y}$, the following equation holds:
\begin{equation}
{\rm Pr}[\mathcal{M}(x)= y]\leq {\rm exp}(\epsilon)\cdot{\rm Pr}[\mathcal{M}(x')= y]
\end{equation}
\end{definition}
A smaller $\epsilon$ guarantees stronger privacy protection because the adversary has lower confidence when trying to distinguish any pair of inputs $x, x'$.

\begin{definition}[Sensitivity~\cite{dwork2006differential}]
For any pair of neighboring inputs $d, d' \in \mathcal{D}$, the sensitivity $\Delta f$ of query function $f(\cdot)$ is defined as follows:
    \begin{equation}
        \Delta f = \max\limits_{d, d'} \left\|f(d) - f(d')\right\|_1,
    \end{equation}
    where the sensitivity $\Delta f$ denotes the maximum change range of function $f(\cdot)$.
\end{definition}

\begin{definition}[Exponential Mechanism~\cite{4389483}]
    Given a score function $q: (\mathcal{D}, y) \rightarrow \mathcal{Y}$, a random algorithm $\mathcal{M}$ satisfies $\epsilon$-differential privacy, if
\begin{equation}
    \mathcal{M}(\mathcal{D},q)=\left\{y:\vert {\rm Pr}[y\in \mathcal{Y}] \varpropto {\rm exp}(\frac{\epsilon(\mathcal{D}, y)}{2\Delta q})\right\}
\end{equation}
where $\Delta q$ is the sensitivity of the score function $q: (\mathcal{D}, y) \rightarrow \mathcal{Y}$.
\end{definition}

\begin{definition}[Constrained domain~\cite{cao2020pglp}]
    We denote $\mathcal{C}^t = \{x_i| {\rm Pr}(x^t = x_i)>0, x_i \in \mathcal{X}\}$ as constrained domain, which indicates a set of possible locations at $t$, where $x^t$ is the user's true location at $t$ and $x_i\in \mathcal{X}$.
\end{definition}

\begin{table*}[t]
\caption{Comparing different gradient inversion attacks in spatiotemporal federated learning. We show the Attack Distance of various attacks on the NYCB dataset, where the local training model employs an LSTM architecture (Attack Distance: $L_2$ distance between true and reconstructed locations}
\centering
\small
\begin{tabular}{|c|c|c|c|c|c|}
    \hline
      Method       &  Initialization &  \makecell{Optimization  terms}   & Model &  Additional strategies & Attack Distance \\
      \hline
      DLG \cite{zhu2019deep}  & Gaussian  &  $l_2$ distance   & LeNet&  -   &  734\\
      \hline
      iDLG \cite{zhao2020idlg}  &  Uniform&\makecell{$l_2$ distance} &LeNet& -  & 690\\
      \hline
      InvGrad \cite{geiping2020inverting} &  Gaussian&\makecell{Cosine similarity}& ResNet&  TV norm  & 2385\\
      \hline
      CPL \cite{wei2020framework} & Geometric & \makecell{$l_2$ distance}&LeNet &   \makecell{label based \\ regularizer} &1329\\
      \hline
      SAPAG \cite{wang2020sapag}& Constant & \makecell{Gaussian kernel \\based function}  &ResNet &  -&  1037 \\
      \hline
      \textbf{ST-GIA} &  ST-based &  \makecell{$l_2$ distance} &LSTM & \makecell{ Mapping, \\ Calibration} & 264 \\
      \hline
      \textbf{ST-GIA+} & ST-based  &  \makecell{$l_2$ distance} &ST models & \makecell{Candidate set, \\ Calibration} & 221\\
      \hline
    \end{tabular}
\label{comparison_state_of_art}
\end{table*}

\section{Spatiotemporal Gradient Inversion Attack}\label{Spatio-temporal Gradient Inversion Attack}
The reconstruction of input data from gradients using gradient inversion attacks has been widely studied. However, existing methods primarily focus on reconstructing training images or text and have not yet been validated in the context of spatiotemporal federated learning. As shown in Table~\ref{comparison_state_of_art}, we conduct a preliminary evaluation of these methods within the spatiotemporal federated learning framework. Our results indicate that while some methods can partially recover real data, the overall success rate remains low. We attribute this limited effectiveness to the lack of consideration for spatiotemporal features in current approaches.
To address these shortcomings, we propose ST-GIA, a novel gradient inversion attack algorithm specifically designed for spatiotemporal data. ST-GIA leverages the unique characteristics of spatiotemporal data to more accurately reconstruct the original locations from gradients. As illustrated in Fig.~\ref{fig: system}, our method consists of three key steps: initialization based on spatiotemporal features, recovering location through gradient matching, and calibration from multiple recoveries.

\subsection{Initialization based on Spatiotemporal Features} 
To reconstruct the spatiotemporal data, we first initialize dummy data, denoted as $(x', y')$, where $x'$ represents the dummy input and $y'$ is the dummy label. Subsequently, we derive the corresponding dummy gradient as follows:
\begin{equation}
    \nabla w^{\prime} \leftarrow \partial \mathcal{L}(F(x^{\prime},w_t),y^{\prime})/\partial w_t.
\end{equation}

There are various strategies for initializing dummy data, each with its own advantages and applications. Among these, the use of random Gaussian noise stands out as the most commonly employed technique for data initialization in image and text recovery tasks.  In addition to Gaussian noise, constant values or random noise sampled from uniform distribution are also presented for data initialization. These alternative approaches can provide different characteristics that may be beneficial depending on the specific context of the recovery task.
Research conducted by Jonas et al.~\cite{geiping2020inverting} has demonstrated that gradient inversion attacks frequently struggle to achieve convergence when initialization is poorly executed. Our experiments in Section V further corroborate this observation, demonstrating that inadequate initialization significantly undermines the effectiveness of attacks in spatiotemporal federated learning.
Considering these insights, we propose that attackers can strategically leverage the inherent characteristics of spatiotemporal data for more effective dummy data initialization. Specifically, an attacker might utilize the reconstructed location from a previous iteration as a foundational reference for initializing the dummy point in the subsequent round. This approach takes advantage of the continuity inherent in user mobility and makes an educated guess, thereby increasing the likelihood of successful data reconstruction. Therefore, for the attack in the $t$-th round, we can initialize the dummy data as follows:

\begin{equation}
    x^{\prime t} \gets x^{\prime t-1}, 
    y^{\prime t} \sim \mathcal{N}(0,1).
\end{equation}
In the case of attacking the first training data round, we continue to employ random Gaussian noise for the initialization of dummy data.

\subsection{Gradient Matching}
The next step involves optimizing the dummy gradient, $\nabla w^{\prime}$, to closely approximate the ground truth gradient, $\nabla w$. To achieve this, we must define a differentiable distance function, $D(\nabla w^{\prime}, \nabla w)$, enabling us to determine the optimal $x'$ and $y'$, denoted as $(x^*, y^*)$, as follows:
\begin{equation}
    (x^*, y^*)= \arg\min\limits_{(x', y')} D(\nabla w^{\prime}, \nabla w).
\end{equation}

\textbf{Distance Function.} We consider the $l_2$ norm (Euclidean distance) as our distance function to measure the difference between $\nabla w^{\prime}$ and $\nabla w$.  The Euclidean distance fits the characteristics of the spatiotemporal data since it is a natural metric of distance between locations.
\begin{equation}
    D(\nabla w^{\prime}, \nabla w) = \left \| \nabla w^{\prime}-\nabla w \right\|_2.
\end{equation}

\textbf{Mapping.} Our research focuses on tasks related to predicting human mobility, allowing us to leverage the road network to significantly enhance the accuracy of our attacks. We assume that all location coordinates are on the road network, which is a reasonable assumption since many human mobility prediction tasks are based on the road network.
To effectively incorporate this assumption into our attack optimization process, we implement a critical step: if the output of an attack iteration results in a location $x'$ that lies outside the boundaries of the road network, we map it to the nearest point within the network. This mapping enhances the effectiveness of our attacks by ensuring that the results align with the actual patterns of human activity.
Ablation experiments presented in Section~\ref{PERFORMANCE EVALUATION} provide compelling evidence that this straightforward operation can significantly improve the success rate of the attacks. By constraining the generated positions to the road network, we facilitate the convergence of the attack model, ensuring that the results of each iteration remain within a plausible and contextually appropriate range.

\begin{algorithm}[t]
\caption{ST-GIA}\label{Algorithm 1}
\begin{algorithmic}[1] 
\REQUIRE{$F(x; w_t)$: Differentiable machine learning model, $\nabla w_t$: model gradients after target trains at round $t$, learning rate $\eta$ for optimizer, $N$: max attack iterations. $T$: global training rounds.}
\ENSURE{reconstructed training data $(x^{*},y^*)$}
\FOR{$t=1$ to T }
\IF{$t=1$}
\STATE Initialize dummy locations $x_0^{\prime} \sim \mathcal{N}(0,1), y_0^{\prime} \sim \mathcal{N}(0,1)$.
\ELSE
\STATE Initialize dummy locations $x_0^{\prime t} \gets x_{N+1}^{\prime t-1}, y_{0}^{\prime} \sim \mathcal{N}(0,1)$.
\ENDIF
\FOR{$i \leftarrow 1$ to $N$}
  \STATE $\nabla w_i^{\prime} \leftarrow \partial \mathcal{L}(F(x_i^{\prime},w_t),y_i^{\prime})/\partial w_t$
    \STATE $\mathbb{D} \leftarrow \left \| \nabla w_i^{\prime} - \nabla w  \right \|^2$
    \STATE $x_{i+1}^{\prime} \leftarrow x_i^{\prime} - \eta\nabla_{x_i^{\prime}}\mathbb{D}_i, y_{i+1}^{\prime} \leftarrow y_i^{\prime} - \eta\nabla_{y_i^{\prime}}\mathbb{D}_i$
    \STATE \textit{Mapping} $(x_{i+1}', y_{i+1}' )$
    \ENDFOR
\ENDFOR
\STATE \textit{Calibration} from multiple recoveries according to Equation \ref{Calibration}.
\STATE \textbf{return} $(x^{*}, y^*)$ 
\label{DLG}
\end{algorithmic}
\end{algorithm}

\subsection{Calibration from Multiple Recoveries} 
To effectively capture the spatiotemporal relationships among data points, each data point participates in multiple training iterations. Consequently, for any given data point, an attacker can generate several reconstruction results by conducting attacks across various training rounds. When the total number of global training rounds $t$ is less than the timesteps $T_s$, each point is reconstructed $t$ times. Conversely, if the number of training rounds exceeds $T_s$, each point undergoes $T_s$ reconstructions.
This iterative process allows attackers to leverage the information gained from multiple reconstructions. By aggregating these results, attackers can achieve a more accurate and reliable estimate of the original data points. Specifically, the mean of all reconstructed positions for a data point can be considered the final attack result:
\begin{equation}
x^{*} =
\begin{cases}
\frac{1}{|T_s|} \sum_{T_s} x', & \mbox{if } t \ge |T_s| \\
\frac{1}{|t|} \sum_{t} x', & \mbox{if } t \leq |T_s|
\end{cases}    
\label{Calibration}
\end{equation}
This approach not only enhances the precision of the reconstructed data but also reduces the influence of outliers that may emerge from individual iterations, thereby increasing the robustness of the attack overall. It can be considered as a group consistency from multiple recoveries that helps get closer to the global optimal point.

\subsection{The Framework of the Algorithm}
Algorithm~\ref{Algorithm 1} demonstrates the proposed ST-GIA. In each global round, we initialize dummy data $(x_0', y_0')$ based on spatiotemporal features in lines 2–6. We obtain the dummy gradient $\nabla w^{\prime}$ corresponding to the dummy input in line 8. We then use Euclidean distance to measure the distance between the dummy and true gradients (line 9). After each attack iteration, we update $(x_i', y_i')$ in line 10. If $(x_i', y_i')$ are not within the range of the road network, then we map $(x_i', y_i')$ to the location on the road network closest to it. When the preset maximum number of iterations $N$ is reached or the dummy data no longer changes, we obtain the preliminary attack result. After obtaining all reconstruction results of a location, we calibrate all preliminary results to obtain the final reconstructed position $(x^{*}, y^*)$.

\begin{figure}[ht]
    \centering
    \includegraphics[width=0.7\linewidth]{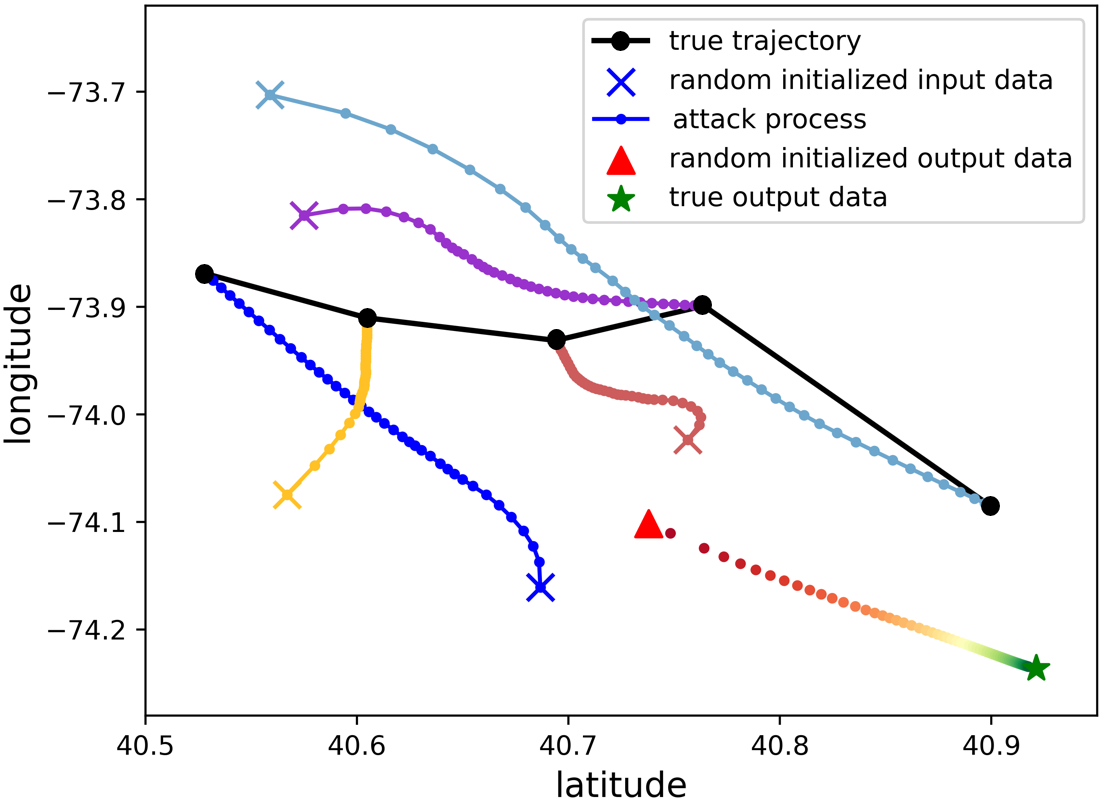}
    \caption{A reconstructed trajectory}
    \label{A reconstructed trajectory.}
\end{figure}

Fig.~\ref{A reconstructed trajectory.} illustrates the performance of an attacker to reconstruct a user’s trajectory utilizing ST-GIA, wherein each reconstructed trajectory point closely approximates the corresponding true trajectory point. We present the attack process on the first-epoch global model, thus employing random initialization. Upon convergence of the attack model, the minimum distance between the reconstructed and true locations is reduced to a mere 1 meter. Moreover, we find that the attack results converge in the direction of the true location, regardless of the initialization location. We conduct multiple experiments under the real-world dataset and found that to be true, independent of the initialization dummy location.


\section{Enhanced ST-GIA with Large Language Models}\label{Spatiotemporal Gradient Inversion Attack+}
Despite the significant advantages and effective attack performance of ST-GIA over traditional gradient inversion methods in the context of spatiotemporal federated learning, our experiments reveal that the attack performance of ST-GIA is closely related to the number of global training rounds. In scenarios with limited global rounds, ST-GIA effectively captures useful information from the gradients, enabling precise reconstruction of the original data. However, as the number of global training rounds increases, the model gradually converges, leading to a decrease in the information available within the gradients and consequently resulting in a marked decline in attack efficacy.

\begin{figure}
        \centering
    \includegraphics[width=0.4\textwidth]{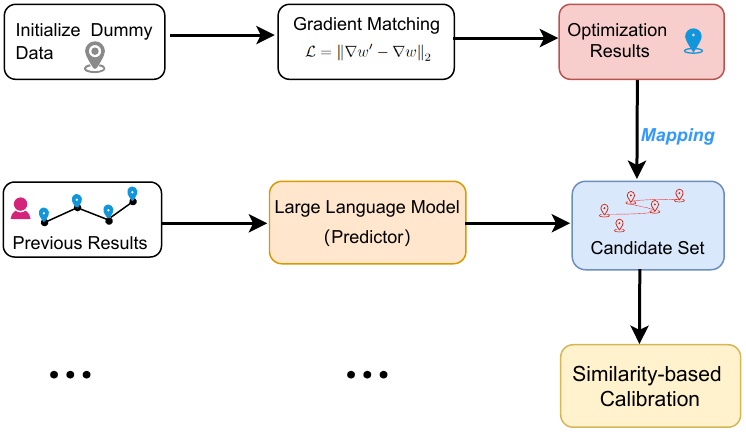}
    \caption{The attack process of ST-GIA+.}
    \label{fig: framework}
\end{figure}

To address this issue and further enhance the performance of ST-GIA in later training rounds, we propose an improved method called ST-GIA+, as shown in Fig.~\ref{fig: framework}. This enhanced approach leverages a large language model to guide the search for potential locations through its powerful generative capabilities, thereby strengthening the ability to reconstruct original data from sparse gradients.
Specifically, the attack process of ST-GIA+ consists of three key steps. First, the attacker uses historical attack results as input for the large language model, which generates a set of possible location candidates for the current time step. Next, the attacker maps the results from the current round to the generated candidate set. This process combines historical attack information with the current round's gradients to narrow the range of possible locations, thereby increasing the accuracy of the attack. Finally, the attacker refines each candidate location in the set by calculating trajectory similarity, ultimately yielding an attack result that aligns most closely with the actual trajectory.

\subsection{Initialization} We employ the same spatiotemporal feature-based initialization method as ST-GIA to generate dummy data. Specifically, we fully leverage the continuity of human mobility by using the reconstruction results from the previous round of attacks as the initial input for the next round. This continuity-based initialization strategy not only effectively reduces the number of iterations required for each attack round but also enhances the precision of the attacks.

\subsection{LLM-based Candidate Set Generation} Large language models (LLMs) have been extensively utilized in human mobility prediction tasks in recent years, demonstrating exceptional performance. To further enhance the effectiveness of our attacks, we employ the LLM-Mob model proposed by Wang et al.~\cite{wang2023would} to generate potential candidate locations for target positions. The LLM-Mob model introduces two key concepts: "historical stay" and "situational stay", which are designed to capture both long-term and short-term dependencies in human mobility. Historical stay reflects an individual's travel patterns over an extended period, while situational stay emphasizes short-term behavioral patterns relevant to the current context. By integrating these two concepts, LLM-Mob can more accurately model the complexity of human mobility. To enhance the model's predictive capabilities further, LLM-Mob employs a context-based prompting mechanism that leverages rich contextual information to provide clearer input for the LLM. This approach enables the model to generate more precise predictions, thus offering a more accurate set of candidate locations for the attacks. Specifically, given the previous reconstruction results:
\begin{equation}
    s_u^*=\{(l_1^*, t_1), (l_2^*, t_2), \dots, (l_{i-1}^*, t_{i-1})\}
\end{equation}
Then, we use a large language model to generate candidates for the new reconstruction $(l_i, t_i)$. We use recall@5 to evaluate prediction performance, meaning five predictions are provided for each location. 

\subsection{Mapping Reconstruction Results} 
\begin{figure}
    \centering
    \includegraphics[width=0.8\linewidth]{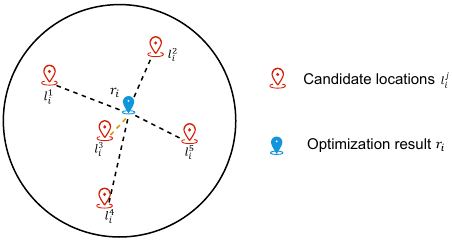}
    \caption{Mapping reconstruction results}
    \label{fig: car}
\end{figure}

We employ the LLM-Mob model as our predictor, generating a candidate set for the next location based on previous attack results. Compared to existing deep neural network-based predictors, LLM-Mob has demonstrated superior accuracy. After generating the candidate set for the next location, we map the attack results to the nearest candidate position. The previously mentioned ST-GIA method improves reconstruction outcomes by mapping attack results to the nearest points on the road network; however, we have observed that the accuracy of the attack significantly increases when the attacker possesses additional prior knowledge. Therefore, after obtaining preliminary attack results by optimizing the distance between the true gradients and the fabricated gradients, we further map these results to the candidate location set generated by the LLM, as illustrated in Fig.~\ref{fig: car}. 

We use the $L_2$ distance between the candidate location set and the attack results as our evaluation metric. Specifically, we select the candidate location that is nearest to the reconstructed location as the final attack result:

\begin{equation}
    x_i^* = \min_{j}\lVert r_i-l_i^j \rVert_2
\end{equation}
where $r_i$ is the reconstruction result obtained after iteration and $l_i^j$ is the $j$-th candidate location by the predictor. 

\subsection{Similarity-based Trajectory Calibration}
In ST-GIA, we directly aggregate the average of all attack results as the final attack result. This method is effective because it combines multiple attack results to improve accuracy. However, directly using the averaged results presents several issues. Specifically, if the attack results of a particular round deviate significantly from the others, it can adversely affect the final results. To address this problem, we propose a similarity-based approach to calibrate the final attack results. Specifically, when the attacker obtains multiple rounds of attack results, they first screen to remove outliers that could negatively impact the final results. Subsequently, the attacker calculates the cosine similarity between any two attacks and compares them to identify a batch of trajectories with the highest similarity. The average location of this batch is then aggregated to produce the final attack results. This similarity-based calibration ensures that the final aggregated result is more robust and less influenced by anomalous data, leading to more accurate and reliable predictions. The similarity is calculated as:
\begin{equation}
    sim(s_i, s_j)=\frac{<s_i,s_j>}{\lVert s_i\rVert, \lVert s_j\rVert}
\end{equation}
where $s_i$ and $s_j$ are arbitrary reconstruction trajectories. 
ST-GIA+ not only retains the advantages of ST-GIA in earlier rounds but also successfully compensates for the decline in attack performance during later rounds by incorporating a large language model and a trajectory similarity correction mechanism. This results in significant improvements in both the success rate and accuracy of the attacks.

\section{Adaptive Privacy-preserving Strategy}
Provable differential privacy may remain the only way to guarantee formal privacy against gradient inversion attacks. We evaluate three differentially private methods on three real-world datasets, as described in Section~\ref{PERFORMANCE EVALUATION}. We find that achieving effective outcomes by directly applying existing differential privacy-based defense methods proves challenging. The reason is that these methods are primarily designed for general-purpose defenses and are not tailored to address gradient inversion attacks specifically. 
To address this gap, we design an adaptive privacy-preserving strategy to mitigate gradient inversion attacks in spatiotemporal federated learning, which mainly consists of adaptive budget allocation and perturbation based on personalized constraint domains. This tailored approach allows for more effective privacy preservation while considering the specific characteristics of the attack.

\subsection{Adaptive Budget Allocation} 
We have observed that the reconstruction error tends to increase as the global model converges. This phenomenon occurs because, with model convergence, the information in the gradients diminishes, and attackers rely on the information leakage from gradients to reconstruct user locations accurately. Consequently, the privacy risks encountered in each training round are not uniform. Traditional methods assume that each training round is equally important, resulting in a uniform distribution of privacy budgets across all rounds. 
However, this uniform allocation method may face challenges in effectively mitigating gradient inversion attacks.
Specifically, training rounds where the model updates are more informative and, therefore, more sensitive might receive insufficient protection, making them prime targets for attackers. Conversely, rounds with less critical update information may be allocated an excess of the privacy budget, resulting in overprotective.

\begin{algorithm}[t]
\small
\caption{Adaptive privacy-preserving strategy.}\label{Algorithm 2}
\begin{algorithmic}[1] 
\REQUIRE{$\epsilon$: the total privacy budget, $T$: max global training rounds, $\mathcal{C}_k^t$: the constraint domain of user $k$ at $t$, $x_k^t$: input data for user $k$ in round $t$.}
\ENSURE{Model parameter $\hat{w_k^t}$.}
\FOR{$t = 1$ to $T$}
    \STATE $\epsilon'=\epsilon-\sum_{i=1}^{t-1}\epsilon_j$
    \STATE $\epsilon_t = exp(-\gamma[t]) \cdot \epsilon'$
    \FOR{each user $k$ \textit{in parallel}}
        \STATE $\hat{x_k^t}=\textbf{PGEM}(\mathcal{C}_k^t, \epsilon_t, x_k^t)$
        \STATE $\hat{w_k^t} \leftarrow LocalUpdate(k, w_{t-1}, t, \hat{x_k^t})$
        \STATE \textbf{return} $\hat{w_k^t}$ to server 
    \ENDFOR
\ENDFOR
\label{Adaptive privacy-preserving strategy}
\end{algorithmic}
\end{algorithm}

To address these challenges, we propose an adaptive privacy budget allocation method. This approach considers the potential risks associated with each training round, thereby achieving a better trade-off between privacy and utility. Specifically, our central idea is that during the training process, users should dynamically adjust the perturbation of their local data in response to the varying importance of different rounds. This assessment of importance is based on the associated attack risk. To measure this risk, we employ two metrics: Attack Distance (AD) and Attack Iteration (AIT). Attack Distance reflects the $L_2$ distance between true and reconstructed locations. A higher attack distance indicates that the attacker has obtained less accurate data, suggesting a lower risk. Conversely, a lower attack distance signifies that the attacker can reconstruct locations more accurately, indicating a higher risk. We also utilize Attack Iteration to assess the cost of the attack. AIT is defined as the number of iterations required for a successful attack. A lower AIT value signifies greater efficiency for the attacker, meaning fewer attempts are needed to breach privacy, thus indicating a higher risk. Conversely, a higher AIT value suggests that the attacker requires more iterations to succeed, indicating a lower risk. By incorporating these two metrics, we can develop a more nuanced understanding of the importance of each training round. The importance of round $t$ can be computed as follows:
\begin{equation}
    \gamma[t]=\frac{1}{\alpha 
    \mathcal{F}_1(AD[t]) + \beta \mathcal{F}_2(AIT[t])}, 
\end{equation}
where the functions $\mathcal{F}_1(\cdot)$ and $\mathcal{F}_2(\cdot)$ denote the effect of AD and AIT on importance, respectively. The parameters $\alpha, \beta$ are weight factors and $\alpha + \beta = 1$.

We aim to provide adaptive protection for the training data. Rounds with a higher risk require intensified security measures, i.e., more noise should be added. Therefore, the proportional function that decides the portion of the remaining budget allocated to the current round can be defined as:
\begin{equation}
    p=exp(-\gamma[t]).
\end{equation}
The exponential function guarantees that $p$ ranges from 0 to 1. The final budget allocated to the current round is
\begin{equation}
    \epsilon_i = p \cdot \epsilon',
\end{equation}
where $\epsilon'$ is the remaining budget $\epsilon' = \epsilon-\sum_{i=1}^{t-1}\epsilon_t$. By dynamically adjusting the allocation of privacy budgets based on the privacy risk of each round, we aim to enhance the overall resilience of the model against gradient inversion attacks while maintaining optimal performance.

\subsection{Personalized Constraint Domain} 
To ensure privacy, each user can define a personalized constraint domain based on their individual requirements. For example, a student who is primarily active only on campus can define her constraint domain as all locations within the entire campus. This approach allows for tailored privacy settings that reflect the unique usage patterns and privacy needs of each user. By defining a constraint domain, users can ensure that the privacy mechanisms in place are aligned with their specific context and activity patterns. This personalized constraint domain acts as a boundary within which the privacy budget allocation and noise addition are managed. As a result, privacy protection becomes more relevant and effective, minimizing unnecessary noise addition outside the user's active area while maximizing privacy within the defined domain. This method not only enhances privacy protection by focusing on areas of higher activity but also optimizes resource utilization by avoiding the indiscriminate application of privacy measures. By concentrating efforts within the user-defined constraint domain, the system can offer a more efficient and user-centric privacy-preserving solution. Personalized constraint domain can be denoted $\mathcal{C}_k^t = \{x_i^k| {\rm Pr}(x_k^t = x_i)>0, x_i \in \mathcal{X}\}$, which indicates a set of possible locations of user $u_k$ at $t$.

We propose an obfuscation mechanism, PGEM, considering each user's personalized constraint domain so that it can output more useful locations. PGEM uses the idea of an exponential mechanism~\cite{4389483} to perturb the true location of user $u_k$. Given the input $x\in \mathcal{X}$, the privacy budget $\epsilon_t$, and the constraint domain $\mathcal{C}_k^t$, PGEM outputs $c\in \mathcal{C}_k^t$ with the following probability:

\begin{equation}
   { \rm Pr[PGEM }(x) = c] =\frac{e^{-\frac{\epsilon_t}{2}d(x, c)}}{\sum_{c\in \mathcal{C}_k^t}e^{-\frac{\epsilon_t}{2}d(x, c)}}{,}
     \label{PGEM}
\end{equation}
where $d(\cdot)$ is the distance metric between two locations. Simply, we can use Dijkstra's algorithm, which means that  $d(x, \cdot)= Dijkstra(\mathcal{C}_k^t, x)$.

\subsection{Adaptive Privacy-preserving Strategy}
We propose an adaptive privacy-preserving mechanism to protect location privacy in spatiotemporal federated learning. Our algorithm is shown in Algorithm~\ref{Algorithm 2}. It consists of three phases: (1) Calculate the privacy budget for the round $t$ in lines 1–3. (2) Obfuscate the input data $x_k^t$ of client $k$ according to Eq.~\ref{PGEM}. (3) Client $k$ utilizes the obfuscated data $\hat{x_k^t}$ for local updates, thereby deriving the model parameters $\hat{w_k^t}$, which are subsequently uploaded to the server.

\section{Experiments}\label{PERFORMANCE EVALUATION}

\subsection{Experimental Setup}
\subsubsection{Datasets} We evaluate the performance of attack and defense strategies on three real-world location datasets: NYCB\footnote[1]{https://www.kaggle.com/datasets/stoney71/new-york-city-transport-statistics}, Tokyo\footnote[2]{https://sites.google.com/site/yangdingqi/home/foursquare-dataset}, and Gowalla\footnote[3]{http://snap.stanford.edu/data/loc-gowalla.html}. The statistics of each dataset are shown in Table~\ref{tab: Basic dataset statistics.}. For each dataset, we randomly select 100 users to participate in federated training and select trajectory points in roughly 10-minute increments.

\subsubsection{Implementation Details of LLM} 
We employ the LLM-Mob model proposed by Wang et al.~\cite{wang2023would} to generate potential candidate locations. The specific large language model utilized in our experiments is GPT-3.5\footnote{https://platform.openai.com/docs/models/gpt-3-5}, which is one of the most advanced and widely used LLMs available through open APIs and offers relatively lower costs. To eliminate randomness in the output, we set the temperature to 0. Additionally, the lengths of historical stays and contextual stays are set to 40 and 5, respectively. These values are empirical and can be adjusted to meet specific needs.

\begin{table}[]
\caption{ Basic dataset statistics.}
    \centering
    \begin{tabular}{cccc}
    \hline
                 &  NYCB   \qquad  &Tokyo \qquad    & Gowalla\\    \hline
     $\#$users      &   1064   &2293     &53008    \\
     $\#$locations  &   5136   &7872     &121944  \\  
     $\#$check-ins  &   147939  & 447571 & 3302414\\ \hline
    \end{tabular}
    \label{tab: Basic dataset statistics.}
    \label{Table 2}
\end{table}

\subsubsection{Metrics} 
We utilize the Attack Distance (AD) to evaluate the effectiveness of the reconstruction attack. AD represents the $L_2$ distance between the reconstructed location and the true location, with a smaller AD indicating a stronger attack capability. Additionally, we employ the Attack Iteration (AIT) to measure the cost of the attack, defined as the number of iterations required for the attack to succeed~\cite{wei2020framework}. To evaluate the effectiveness of the defense strategy, we use the top-$k$ recall rate (recall@5) as an indicator of prediction performance. For each experiment, we execute the methods 100 times and calculate the average results. 

\subsubsection{Local Models} To evaluate the impact of gradient inversion attacks on different local models, we conduct attack experiments using four distinct local models, including LSTM~\cite{hochreiter1997long}, PMF~\cite{feng2020pmf}, DeepMove~\cite{feng2018deepmove}, and STAN~\cite{luo2021stan}.
\begin{itemize}
    \item LSTM. A classic sequence modeling neural architecture that has shown great performance for the next location prediction.
    \item PMF.  A privacy-preserving mobility prediction framework via federated learning, which employs a local LSTM model and utilizes group optimization methods for training.
    \item DeepMove. A framework consisting of two distinct recurrent networks, one for learning the periodicity from historical visits and the other for extracting transition patterns from the current trajectory.
    \item STAN.  A framework uses a bi-layer attention architecture that first aggregates spatiotemporal correlation within user trajectory and then recalls the target with consideration of personalized item frequency.
\end{itemize}

\subsubsection{Attack Methods} In addition to ST-GIA and ST-GIA+, we compare another five attacks that can be applied to spatiotemporal federated learning directly or with simple modifications, including DLG~\cite{zhu2019deep}, iDLG~\cite{zhao2020idlg}, InvGrad~\cite{geiping2020inverting}, CPL~\cite{wei2020framework}, and SAPAG~\cite{wang2020sapag}. See Table~\ref{comparison_state_of_art} for details.

\begin{itemize}
    \item DLG. The images and labels are inferred by minimizing the original gradients and dummy gradients from randomly generated noise data. They use the L-BFGS optimizer for faster convergence.
    \item iDLG.  Extracts ground-truth labels from the gradients. It empirically shows the advantage over DLG while the label extraction only works in single image restoration.
    \item InvGrad. By using cosine distance to measure the gradient difference and using total variation as the regularization for image denoising, the quality of reconstructed images is greatly improved.
    \item SAPAG. A more powerful label leakage attack can be applied to both image classification and speech recognition tasks. The labels can still be revealed after multiple local iterations.
    \item CPL. A framework for evaluating gradient inversion attacks that employs label regularization to enhance the stability of attack optimization.
\end{itemize}

\subsubsection{Defense Strategies} We compare our methods with three defense approaches based on differential privacy, including DPSGD~\cite{abadi2016deep}, GeoI~\cite{andres2013geo}, and GeoGI~\cite{takagi2020geo}. They are widely used to protect privacy in spatiotemporal federated learning.
\begin{itemize}
    \item DPSGD is an extension over the popular stochastic gradient descent algorithm that offers theoretical privacy guarantees. It is a generalized method for training neural networks on sensitive data that modifies vanilla SGD by clipping the gradient computed over each client, followed by the injection of noise into the clipped gradients.
    \item GeoI is a formal notion of location privacy. A mechanism that achieves it guarantees the indistinguishability of a true location from other locations by perturbing the raw data to some extent against any adversary. 
    \item GeoGI is a relaxed version of GeoI that guarantees the indistinguishability of a true location on road networks.
\end{itemize}

\begin{table*}[thbp]
\caption{\small Comparison of different attack methods (NYCB dataset).}
\centering
\renewcommand\arraystretch{1.1}
 {
\small{
\begin{tabular}{|c|p{1.2cm}|p{1.2cm}|p{1.2cm}|p{1.2cm}|p{1.2cm}|p{1.2cm}|}
\hline

\multicolumn{1}{|c|}{\diagbox{Model}{Round}}  &1   & 10  & 20  & 30   & 40 & 50 \\ \hline
\multirow{1}{*}{DLG}     &  43 &      193  &     684   &      1094    &    1878 &  3070   \\ \cline{2-7}   \hline
\multirow{1}{*}{iDLG}    &  36   &      174    &    626   &      1022  & 1642     & 2794   \\ \hline
\multirow{1}{*}{InvGrad}  &  157  &       560      &    1094   &    2118      &  4302   &   9975     \\ \hline
\multirow{1}{*}{CPL}      & 99 &    218   &  788    &      1223       &   2294 &  3039             \\ \hline
\multirow{1}{*}{SAPAG}    & 96  &     237     &   835     &     1461     &   2769     &  4956 \\ \hline   
\multirow{1}{*}{\textbf{ST-GIA}} & 17   &      65     &   217    &     569      &  967      &  1563 \\ \hline
\multirow{1}{*}{\textbf{ST-GIA+GPT3.5}} & 17   &   61     &   203    & 537      &  862      &  959 \\ \hline   
\end{tabular}
}}
\label{table: global comparion1}
\end{table*}

\begin{table*}[thbp]
\renewcommand\arraystretch{1.1}
\caption{\small Comparison of different attack methods (Gowalla dataset).}
\centering
 {
\small{
\begin{tabular}{|c|p{1.2cm}|p{1.2cm}|p{1.2cm}|p{1.2cm}|p{1.2cm}|p{1.2cm}|p{1.2cm}|}
\hline

\multicolumn{1}{|c|}{\diagbox{Model}{Round}}  & 1   & 10  & 20  & 30   & 40 & 50 \\ \hline
\multirow{1}{*}{DLG}        &  68   &     344   &     914   &    1591      &  2295   &   3130  \\ \cline{2-7}   \hline
\multirow{1}{*}{iDLG}      &    59     &  338        &   879    &    1492        & 1964 &   2609 \\ \hline
\multirow{1}{*}{InvGrad}      &   201   &  629         &    1543     &     2950     &   5640  &  7345      \\ \hline
\multirow{1}{*}{CPL}       &  77    &    187   &   944   &       1627      &  2314  &    3548           \\ \hline
\multirow{1}{*}{SAPAG}     &   63  &      236    & 969       &       1443   &   2596     & 3656  \\ \hline   
\multirow{1}{*}{\textbf{ST-GIA}}       &   22  &      107     &    263   &      751     &   991     &  1693 \\ \hline 
\multirow{1}{*}{\textbf{ST-GIA+GPT3.5}}      &  22   &       94    &   224    &    664       & 945       &  1247 \\ \hline           
\end{tabular}
}}
\label{table: global comparion2}
\end{table*}

\subsection{Results on Attack Methods}

\textbf{Comparison of different attack methods.}
We first compare the performance of different attacks on the real datasets in Tables~\ref{table: global comparion1} and~\ref{table: global comparion2}. These attacks are executed across various global training rounds, and their attack distances (AD) are reported, reflecting the distance between the reconstructed locations and the true locations. A smaller attack distance indicates stronger attack capability.
We observe that as the number of global training rounds increases, the attack distances for all methods tend to rise. This trend is attributed to the reliance of these attacks on gradient leakage information to reconstruct user locations. However, as the global model converges, the amount of leaked gradient information diminishes, resulting in increased reconstruction errors.
Furthermore, we find that ST-GIA consistently outperforms existing attacks in the same rounds. This superior performance is attributed to our customized attack design tailored for spatiotemporal data, validating the effectiveness of the proposed method. Additionally, ST-GIA+ demonstrates better performance than ST-GIA, which can be attributed to the utilization of a large language model to generate candidate locations, providing the attacker with stronger prior knowledge. In contrast, InvGrad consistently performs the worst. This poor performance is due to its use of cosine loss as the distance function, leading to a slower convergence rate. In our experiments, InvGrad typically requires over 20,000 iterations to achieve relatively accurate results, while we limit the maximum number of attack rounds to 200. Due to space constraints, we only present results from two datasets; similar trends are observed in the Tokyo dataset.

\begin{figure}[htbp] 
  \centering           
  \subfigure[Results on NYCB dataset] {\includegraphics[width=4cm]{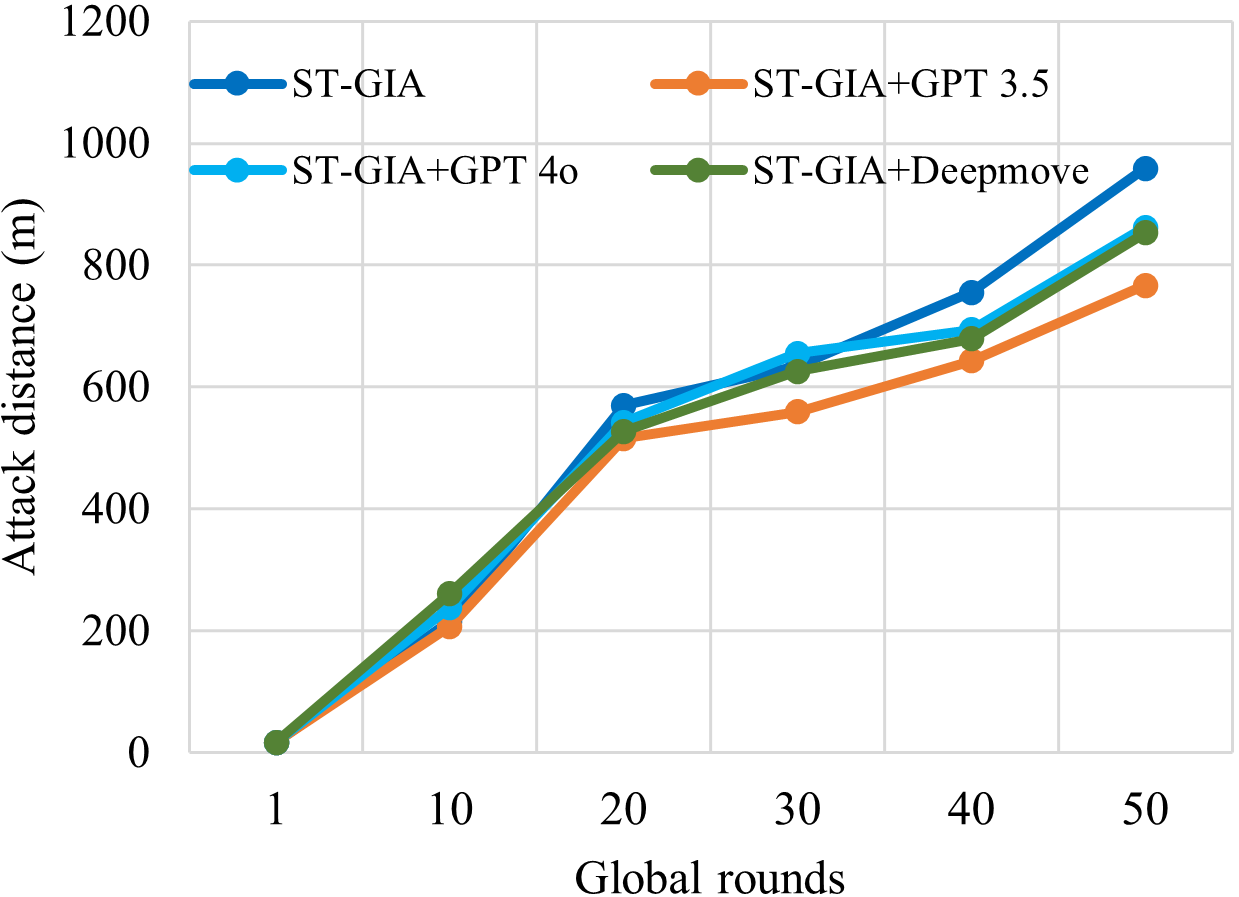}}
  \hspace{0.3cm}
  \subfigure[Results on Tokyo dataset] {\includegraphics[width=4cm]{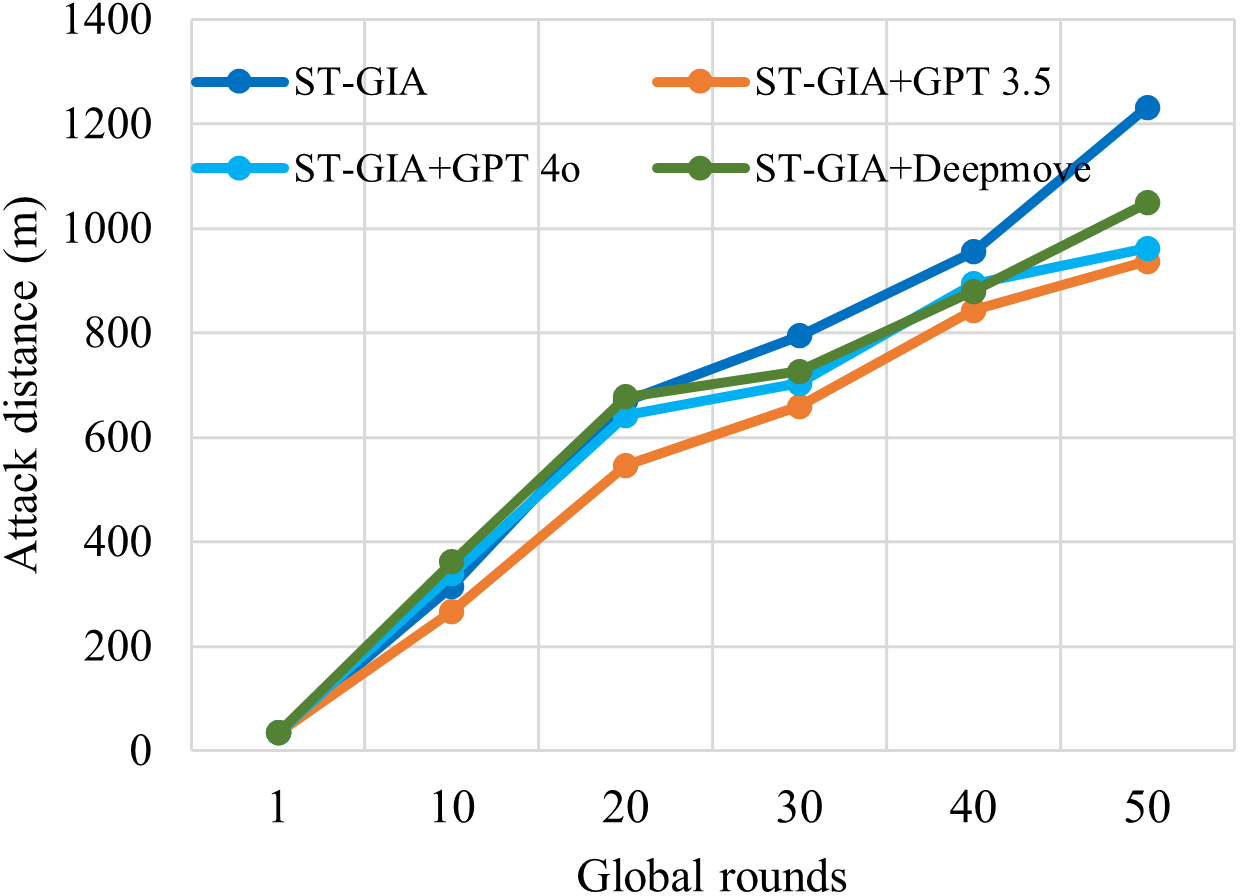}}
  \caption{The impact of different auxiliary predictors.}    
  \label{impact of different auxiliary predictors}    
\end{figure}

\textbf{Comparing the impact of different auxiliary predictors on attack results.}
We compare the impact of different auxiliary predictors on attack results, including large language models and classical predictive models. As shown in Fig.~\ref{impact of different auxiliary predictors}, we find that all auxiliary predictors enhance the attack results to some extent. This improvement is attributed to our enhanced attack method, ST-GIA+, which utilizes pre-trained models to generate candidate sets. Instead of relying solely on gradient matching to derive attack results, attackers identify potential locations that best match the optimized results from the candidate set.
We observe that when the number of global rounds is 1, none of the predictors improve attack accuracy. This limitation arises because auxiliary predictors require previously obtained attack results as input to generate candidate sets.
Furthermore, when comparing ChatGPT 3.5 and ChatGPT 4o, we find that designing appropriate prompts for large language models is crucial. Although ChatGPT 4o is more powerful, it does not demonstrate greater effectiveness in enhancing attacks compared to ChatGPT 3.5 in our experiments. This difference may be attributed to our auxiliary predictors directly utilizing prompts from the LLM-Mob model~\cite{wang2023would}, which is specifically designed for ChatGPT 3.5. OpenAI constantly updates the GPT model family, resulting in performance drift of the newest models. Therefore, prompts that perform well on older models may not be applicable to newer ones, necessitating additional work in prompt engineering.

\begin{figure}[htbp] 
  \centering           
  \subfigure[Results on NYCB dataset] {\includegraphics[width=4cm]{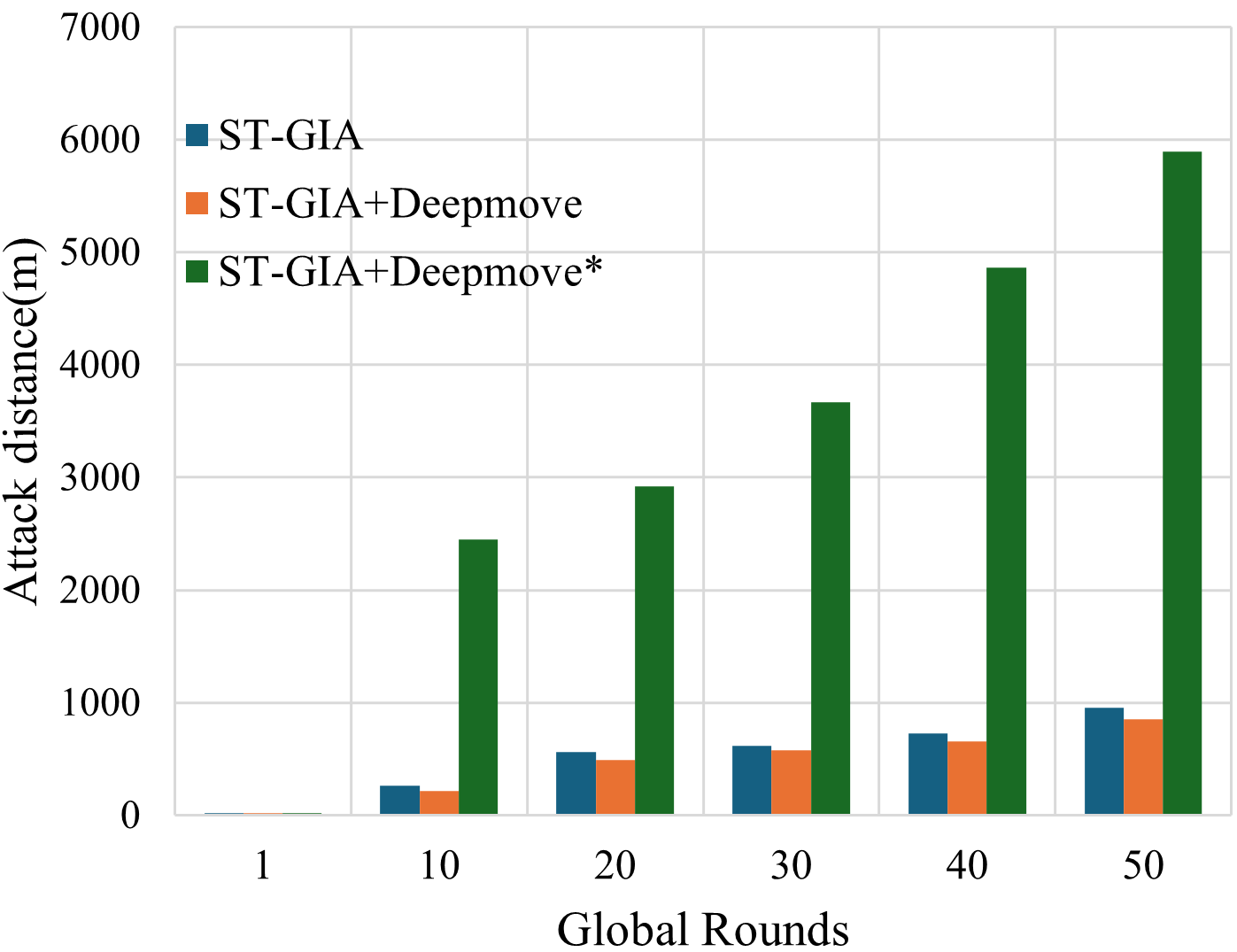}}
  \hspace{0.3cm}
  \subfigure[Results on Tokyo dataset] {\includegraphics[width=4cm]{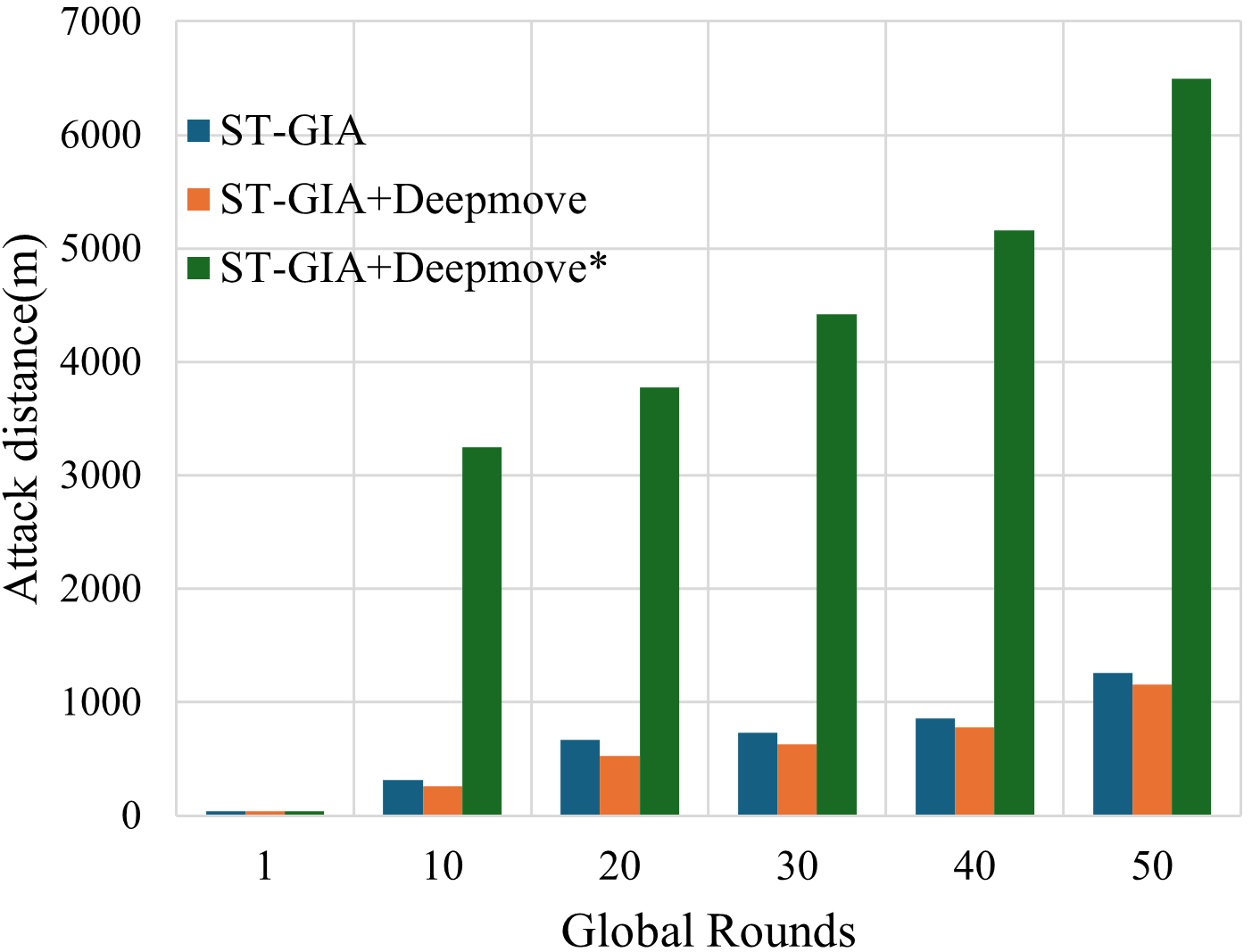}}
  \caption{The impact of predictors trained from different datasets.}    
  \label{impact of different datasets}    
\end{figure}

\textbf{Comparing predictors trained from different datasets.} We also evaluate the impact of predictors trained on different datasets on the attack results. As shown in Fig.~\ref{impact of different datasets}, we find that predictors trained using the same dataset as the global model better enhance the attack results. For example, we compare the impact of predictors trained using different datasets but the same model (DeepMove and DeepMove*) on the attack results. In this experiment, DeepMove uses a mobile app dataset, while DeepMove* uses the same NYCB dataset as the global training model. We find that DeepMove* enhances the attack results more effectively. We attribute this phenomenon to the differing mobility patterns in the datasets, despite being trained similarly. The mobile application dataset collects more data from a person's cell phone, especially the trajectory data generated during walking. In contrast, the NYCB dataset captures bus trajectories, which are more regular due to established routes. Consequently, although the models are the same, the success rate of predictions differs significantly. Therefore, the ability to assist the attacker varies, and models trained on different datasets may even have adverse effects on the attack results.

\begin{figure*}[htbp] 
  \centering           
  \subfigure[Initialization] {\includegraphics[width=5cm]{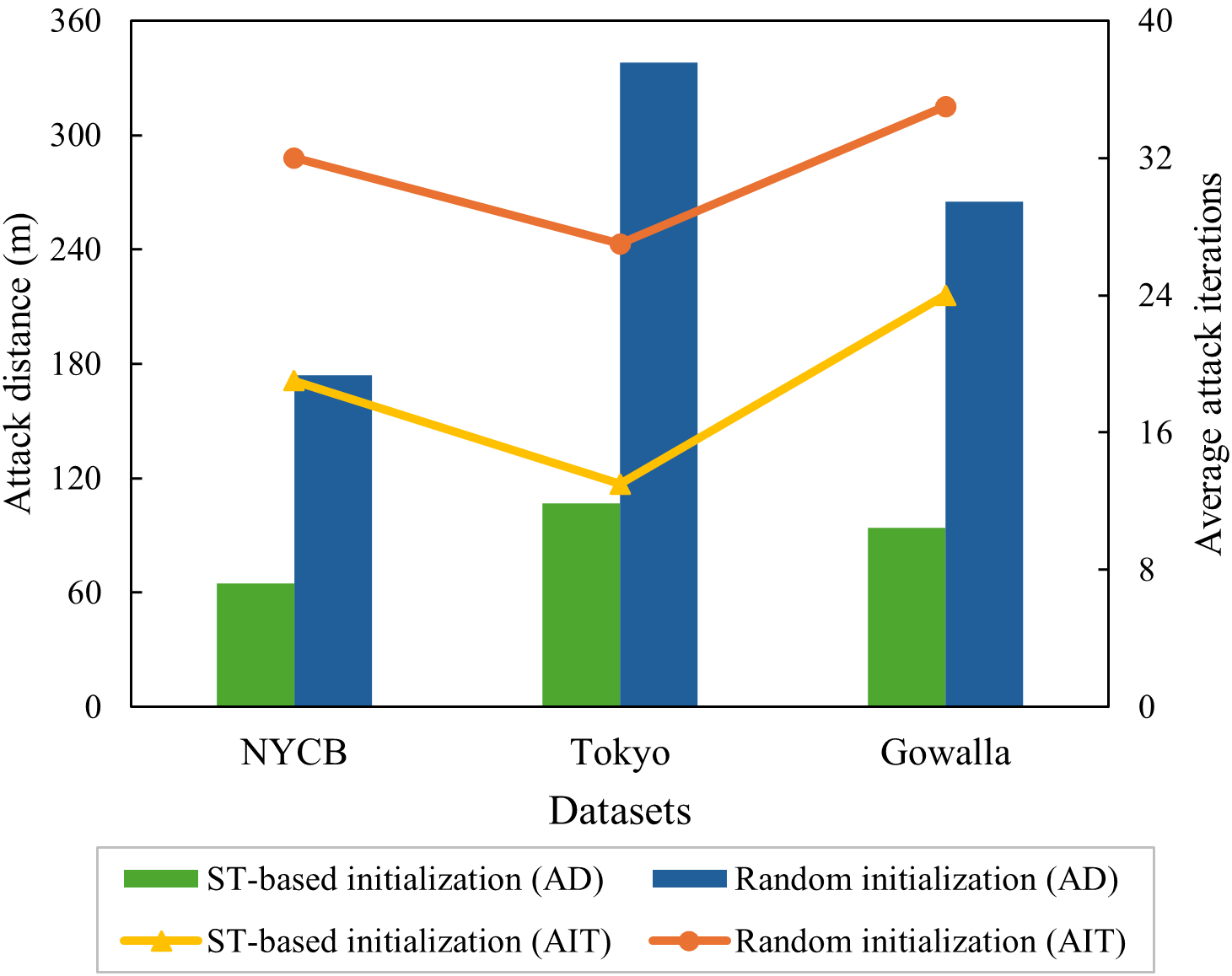}}
  \hspace{0.5cm}
    \subfigure[Mapping] {\includegraphics[width=5cm]{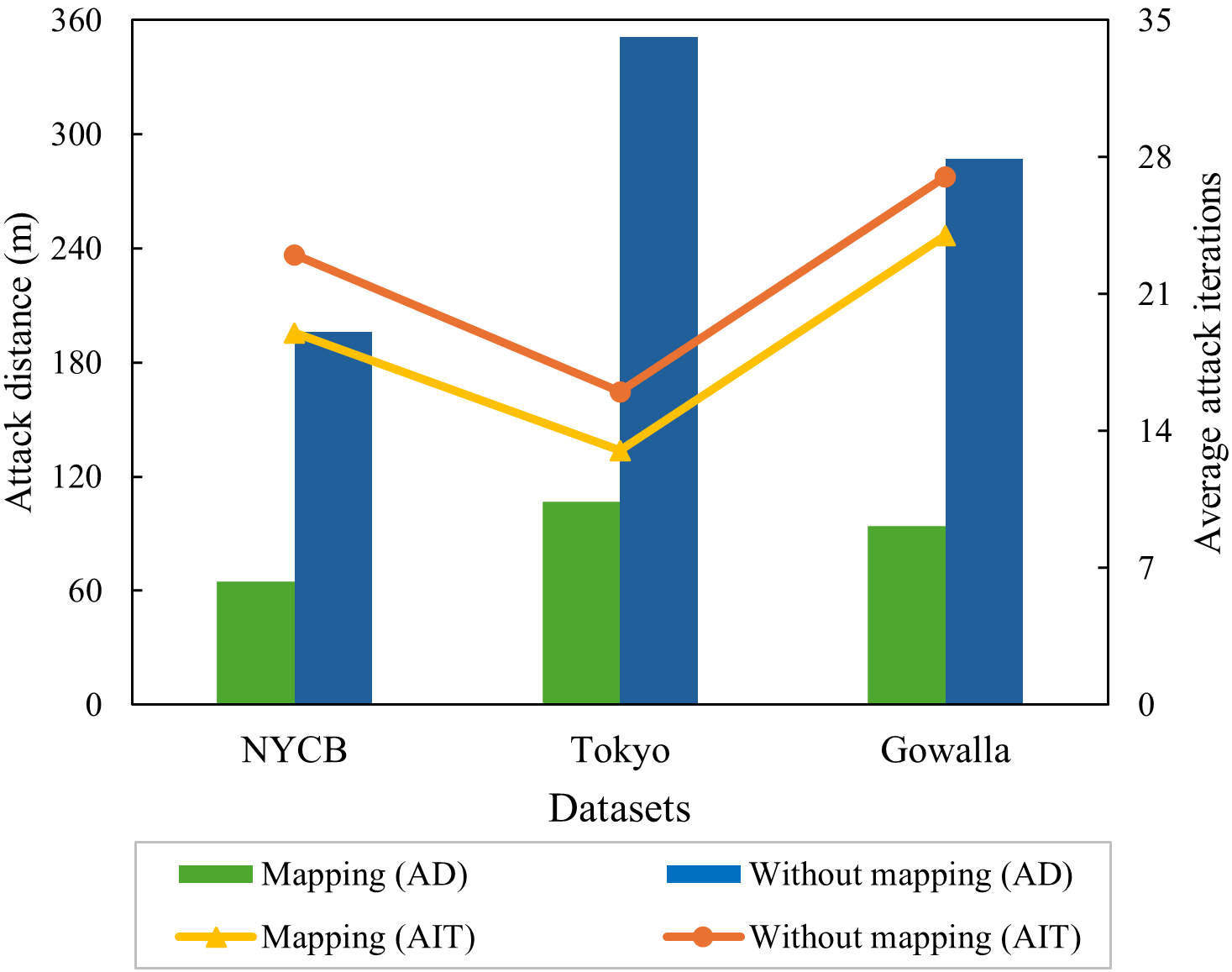}}
    \hspace{0.5cm}
   \subfigure[Calibration] {\includegraphics[width=5cm]{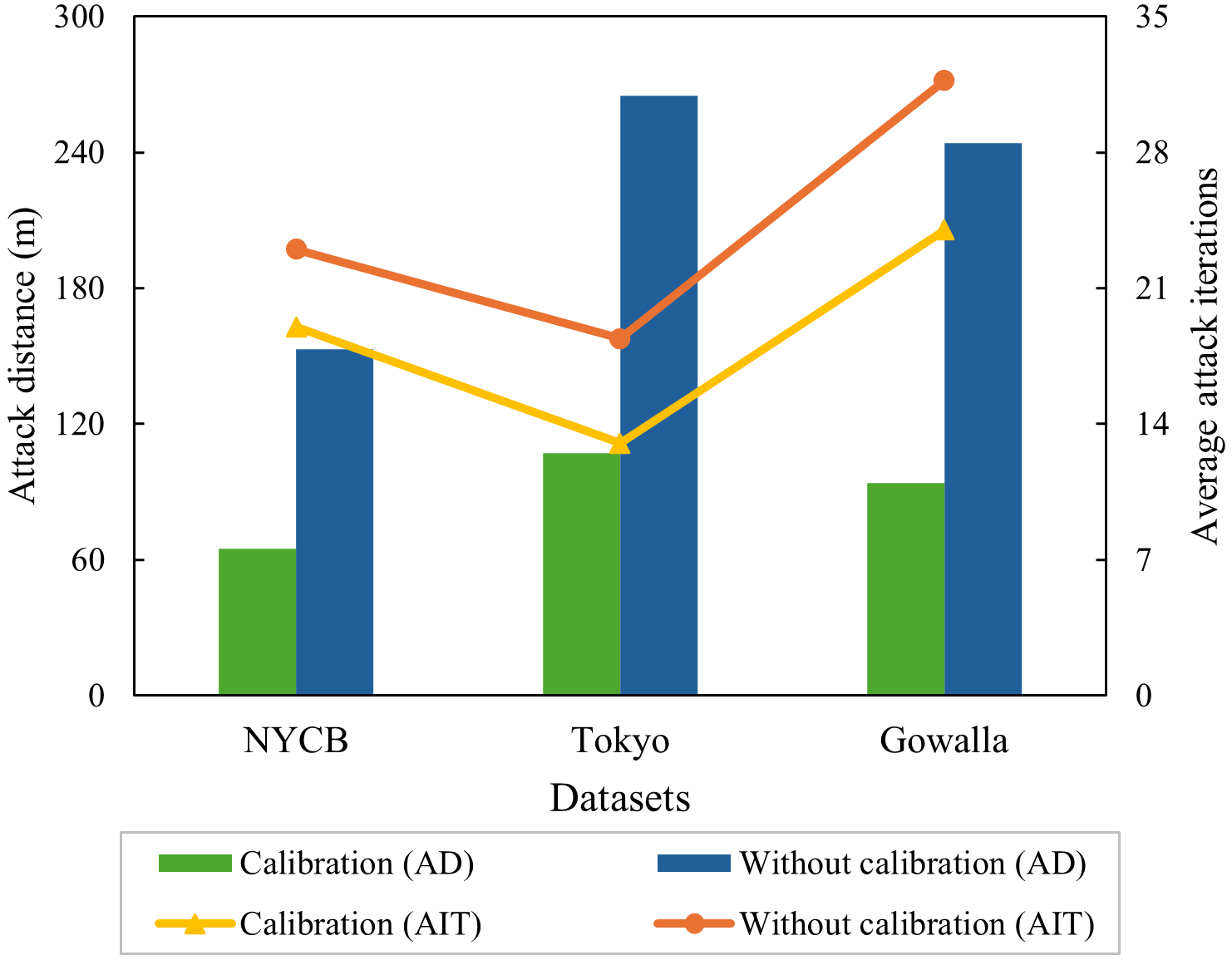}}

  \caption{Ablation studies.}    
  \label{Ablation experiment}    
\end{figure*}
\begin{figure}[htbp] 
  \centering           
  \subfigure[Results on NYCB dataset] {\includegraphics[width=4cm]{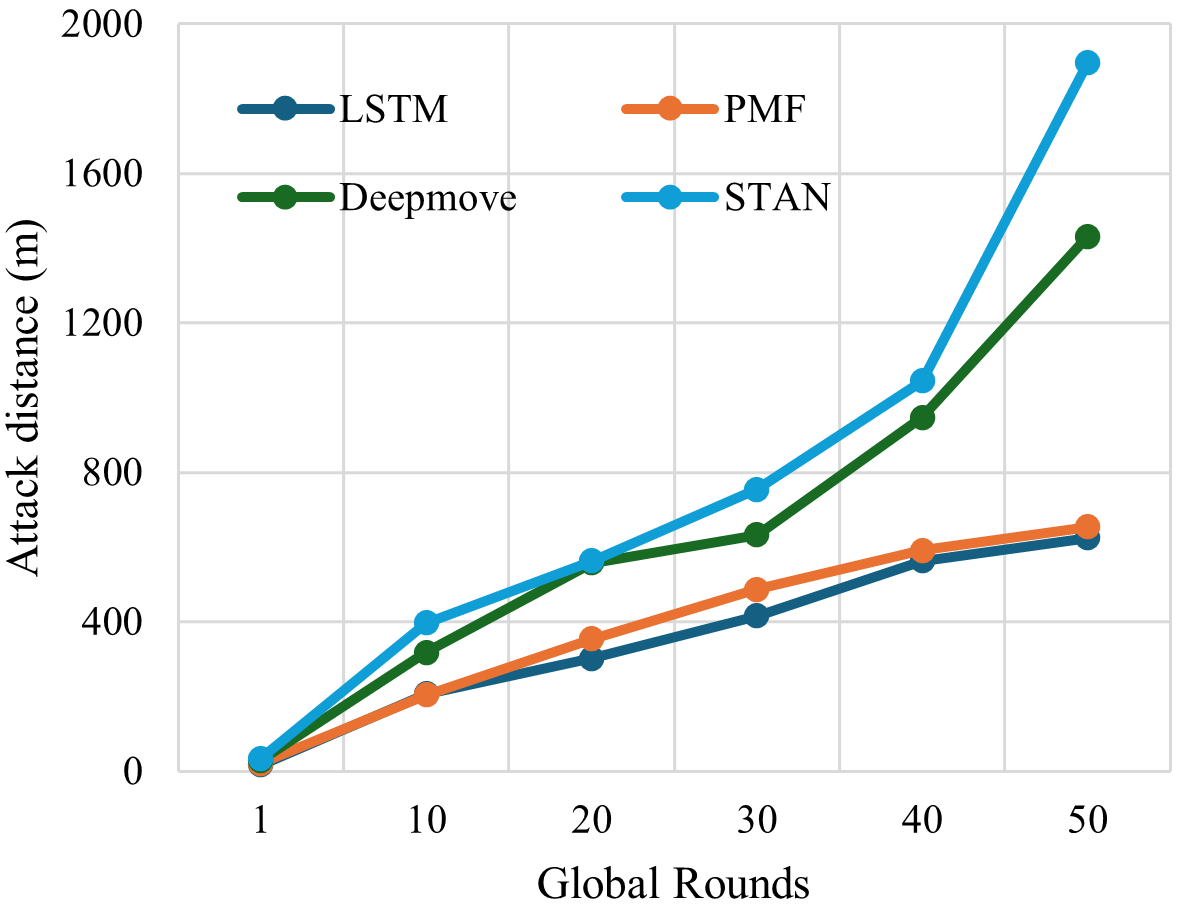}}
  \hspace{0.3cm}
  \subfigure[Results on Tokyo dataset] {\includegraphics[width=4cm]{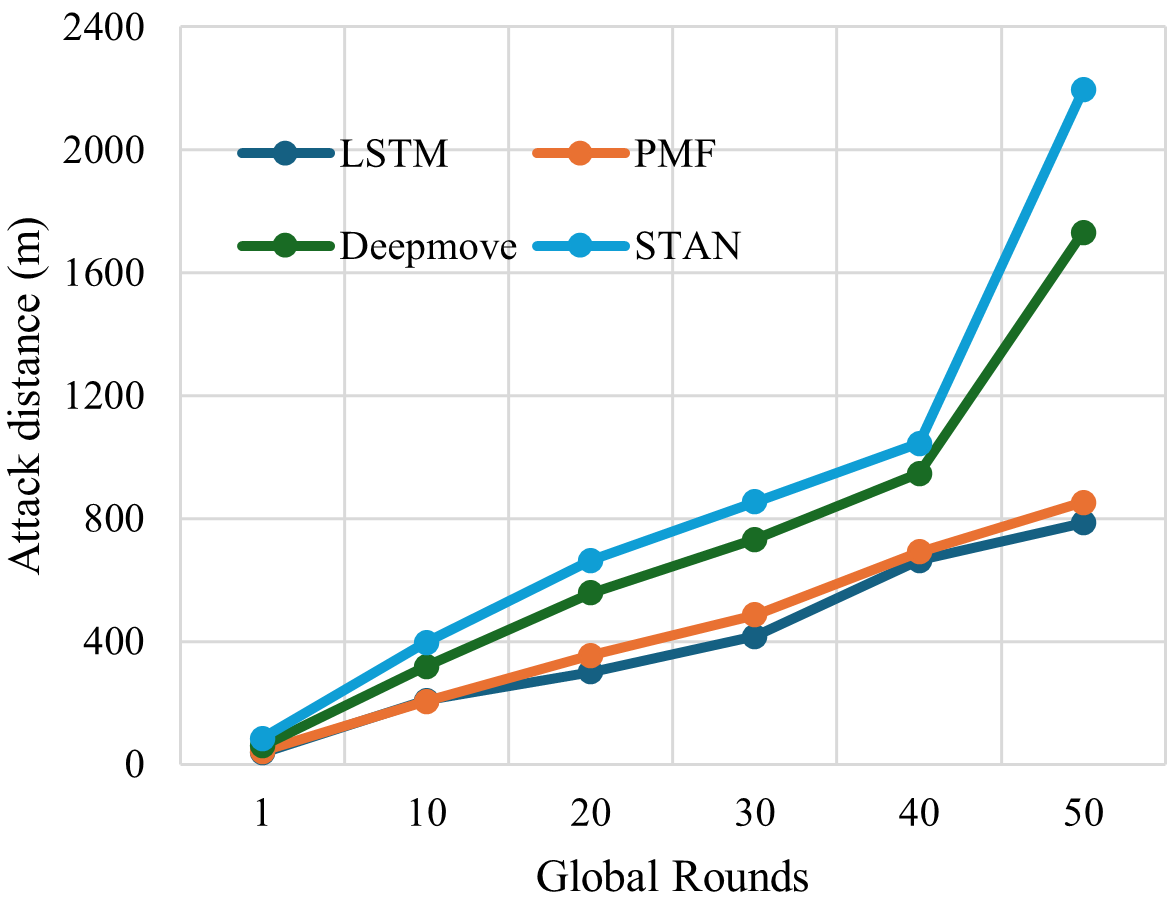}}
  \caption{The impact of different local models on attack results.}    
  \label{impact of different models}    
\end{figure}

\textbf{Comparing the impact of different local models on attack results.} We also conduct experiments to evaluate the performance of different local models in the face of gradient inversion attacks. As shown in Fig.~\ref{impact of different models}, we compare four distinct models: LSTM, PMF, DeepMove, and STAN. In this experiment, we utilize ChatGPT 3.5 as the auxiliary predictor. Our findings indicate that DeepMove and STAN inherently exhibit stronger resilience against gradient inversion attacks compared to LSTM and PMF. For instance, when comparing the LSTM model to the DeepMove model, it is evident that the attack results for the DeepMove model are less favorable under the same number of global rounds. This phenomenon is attributed to the attention-based DeepMove model converging more rapidly than the simpler LSTM model. Consequently, at the same number of rounds, the gradients of the DeepMove model contain less information. Furthermore, due to its faster convergence, the DeepMove model requires less data from users during training, inadvertently reducing the risk of privacy leakage from user data.

\subsection{Ablation Studies} 
As introduced in Section~\ref{Spatiotemporal Gradient Inversion Attack+}, ST-GIA+ enhances the effectiveness of the attack through three key components: initialization based on spatiotemporal features, mapping based on the auxiliary predictor, and trajectory similarity-based calibration. In this experiment, we evaluate the impact of these three components on the attack results.

\textbf{Impact of initialization.}
As shown in Fig.~\ref{Ablation experiment}(a), to evaluate the impact of the initialization component of the proposed attack method on the results, we conduct experiments on three datasets. We initialize the dummy inputs in each round using random and spatiotemporal feature-based initialization methods while keeping other conditions constant. We observe that, compared to random initialization, the spatiotemporal feature-based initialization significantly reduces both the attack distance and the number of attack iterations. Consequently, the spatiotemporal feature-based initialization method enhances the attacker's performance. This improvement is attributed to the fact that poor initialization can hinder the convergence of the attack model and potentially increase associated overhead. In contrast, our spatiotemporal feature-based initialization method leverages the continuity of user mobility to make informed guesses, ensuring that our initialization method is closer to the true locations at the start of each attack round.

\begin{table*}[ht]
 \centering
\caption{ The attack distance of various defense strategies under different privacy budgets $\epsilon$. (Left: NYCB dataset, Middle: Gowalla dataset Right: Tokyo dataset)}
\begin{minipage}[c]{0.30\linewidth}
    \centering
    \setlength{\tabcolsep}{1.5mm}{\scalebox{1}{
    \begin{tabular}{cccccc}
    \hline 
    $\epsilon$       &  1    &5  &10  & 20 & 50 \\  \hline
     DPSGD    &   4960   & 1916    & 1523 &840 &251  \\
     GeoI   &   3421   & 1647  & 1362 & 671 &154\\  
     GeoGI  & 2653   &1820  &1124 &542&78\\ 
     Ours    & 1586 &1064   & 953   & 241 &  65\\ \hline
    \end{tabular}
    }}
\end{minipage} 
\begin{minipage}[c]{0.30\linewidth}
    	\setlength{\tabcolsep}{1.5mm} {\scalebox{1}{
    \begin{tabular}{cccccc}
    \hline 
      $\epsilon$       &  1    &5  &10  & 20 & 50 \\  \hline
     DPSGD    &  6944    &2764  &1848  & 997 & 378 \\
     GeoI   &   4820   & 2351  & 1671 & 816 &257\\  
     GeoGI  & 3711   &1967 &1416 &637 &129\\ 
     Ours    & 2304 &1640   & 1165   & 364 & 94 \\ \hline
    \end{tabular}
    }
    }
   
\end{minipage}
\begin{minipage}[c]{0.30\linewidth}
    	\setlength{\tabcolsep}{1.5mm} {\scalebox{1}{
    \begin{tabular}{cccccc}
    \hline 
      $\epsilon$       &  1    &5  &10  & 20 & 50 \\  \hline
     DPSGD    &  5987    &2386  &1659  & 921 & 367 \\
     GeoI   &   3357  & 2047  & 1453 & 784 &219\\  
     GeoGI  & 2349   &1542  &1226 &557 &136\\ 
     Ours    & 1796 &1461   & 1017  & 306 & 88 \\ \hline
    \end{tabular}
    }
    }
    \label{different privacy budgets}
\end{minipage}

\end{table*}

\textbf{Impact of mapping.} 
Fig.~\ref{Ablation experiment}(b) illustrates the impact of the mapping component in our proposed ST-GIA+ method on attack effectiveness. We utilize ChatGPT 3.5 as the predictor to generate the candidate location set required for mapping. The results indicate that the mapping operation significantly enhances the accuracy of the attacks. This improvement arises from using an effective predictor that assists attackers in generating a set of candidate target locations, thereby ensuring that each attack result falls within a reasonable range through the mapping process. Additionally, this operation reduces the number of iterations required for the attack. This reduction occurs due to the introduction of a spatiotemporal feature-based initialization method, which yields more precise results for each attack round when combined with the mapping operation. More accurate reconstruction results lead to initialization positions closer to the true locations in subsequent rounds, effectively decreasing the number of iterations needed.

\textbf{Impact of calibration.} We also evaluate the impact of the calibration component on the attack results in the proposed attack algorithm. Fig.~\ref{Ablation experiment}(c) shows that calibrating the attack results from multiple reconstructions can enhance attack accuracy to some extent. This indicates that the calibration process improves the attack outcomes, making them more precise. Additionally, it is noteworthy that the calibration process also reduces the number of iterations required for the attack. This reduction occurs because calibration provides more accurate reconstruction positions, further optimizing the initialization results for the subsequent round and thereby decreasing the overall number of iterations needed for the attack. In summary, incorporating a calibration component into the attack algorithm effectively enhances accuracy. This improvement can be entirely attributed to the optimization of attack results during the post-processing stage, making it a crucial addition to the overall attack strategy.

\subsection{Results on Defense Strategies}
In this set of experiments, we compare the performance of different defense methods against gradient inversion attacks. 

\textbf{Comparing different defense strategies.} 
Table~\ref{different privacy budgets} presents the attack distance of various defense strategies under different privacy budgets $\epsilon$ on three real-world datasets. The data indicate that in all cases, the gradient inversion attack is significantly mitigated when sufficient noise is added, albeit at the expense of some accuracy. This finding indicates that all evaluated defense strategies can mitigate gradient inversion attacks to some extent. Additionally, the results demonstrate that for most cases with the same privacy budget, the proposed defense strategy exhibits a higher attack distance, indicating superior resistance to gradient inversion attacks compared to other strategies. These outcomes validate the effectiveness of the proposed defense strategy in enhancing the robustness of models against such attacks. 

\begin{figure}[htbp] 
  \centering           
  \subfigure[Results on NYCB dataset] {\includegraphics[width=4cm]{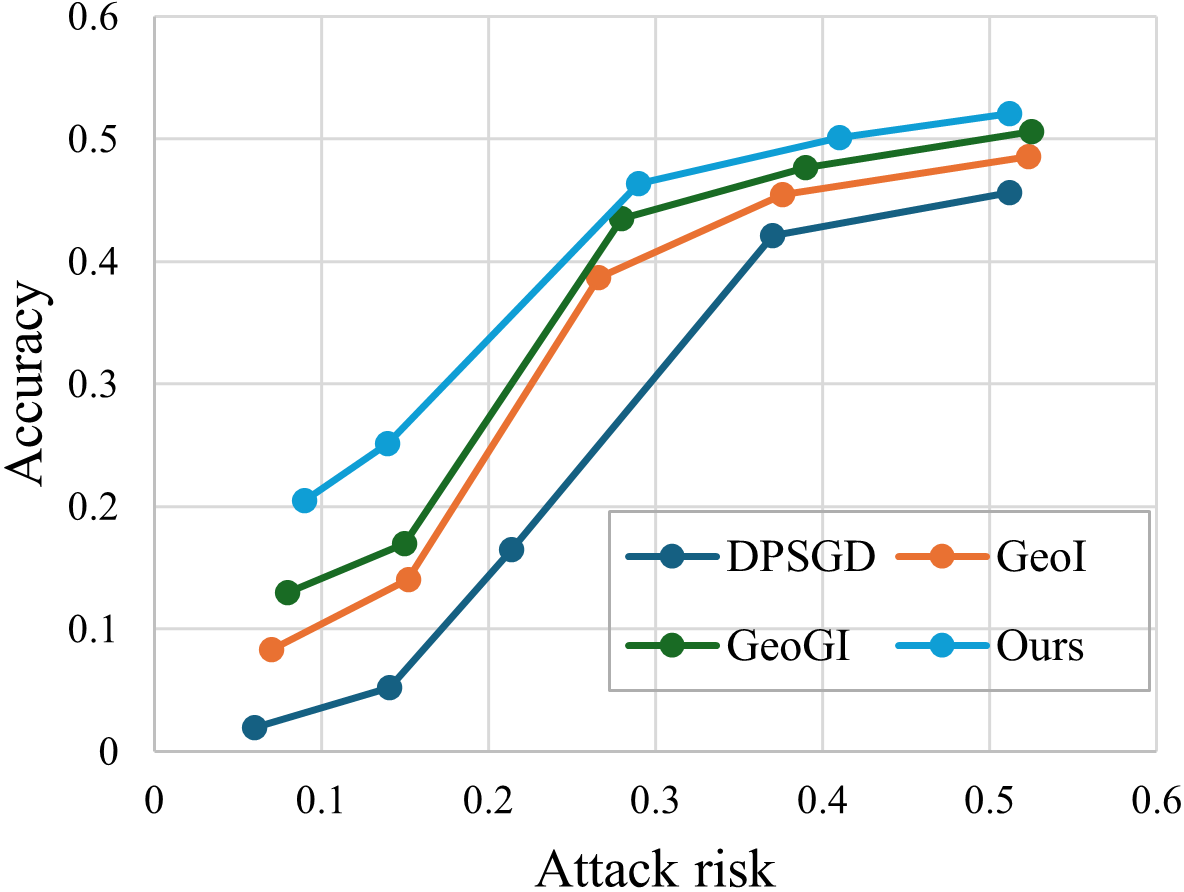}} 
  \subfigure[Results on Gowalla dataset] {\includegraphics[width=4cm]{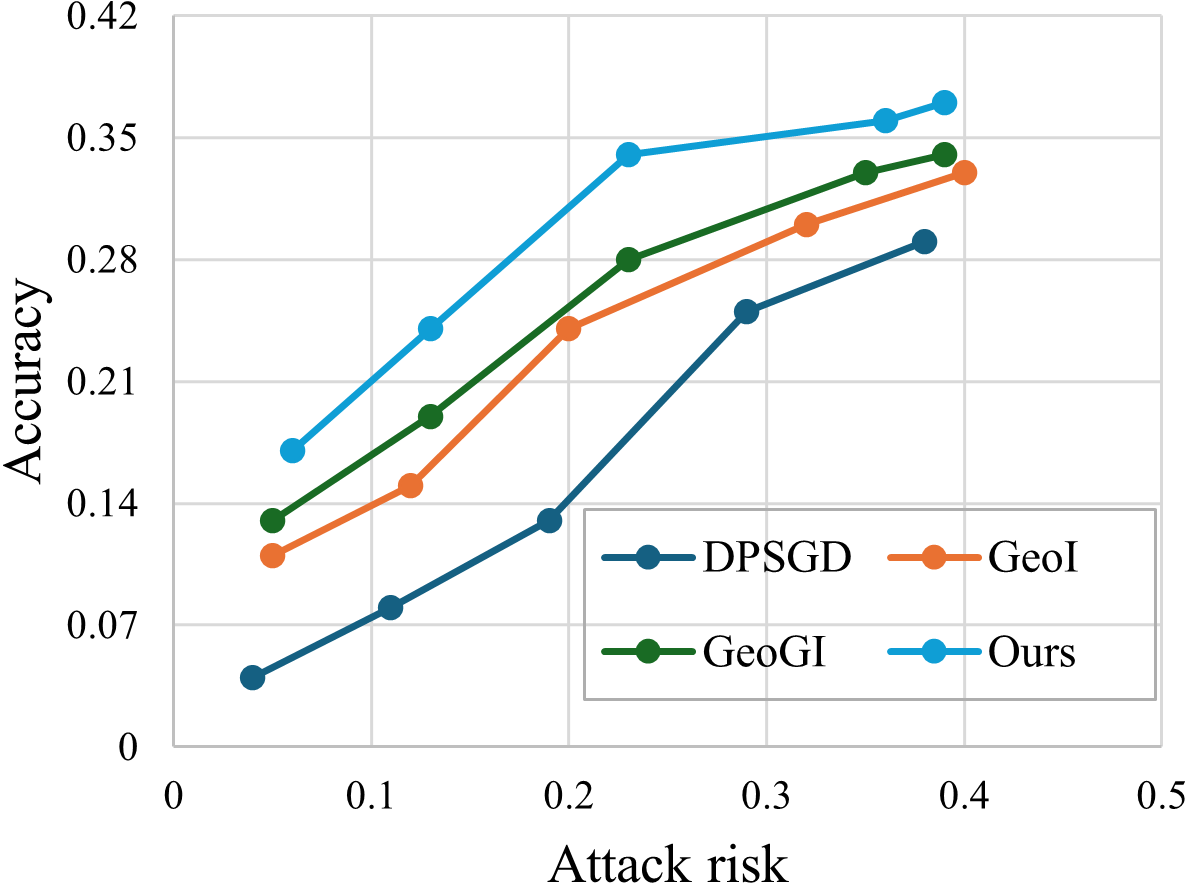}}
  \caption{The relationship between prediction accuracy and attack risk.}    
  \label{Privacy-utility trade-off}    
\end{figure}
\textbf{The trade-off between attack risk and prediction accuracy.} 
To compare different defense strategies, we present the relationship between model prediction accuracy and attack risk in Fig.~\ref{Privacy-utility trade-off}. We define an attack risk as the occurrence of location leakage, indicated by a distance of less than 500 meters between the reconstructed and true locations. Thus, the attack risk can be formulated as follows: $Attack\quad risk=\frac{\left | \{l | l_{attack} < 500m\} \right |}{|L_{total}|}$, where $L_{total}$ represents the set of all locations involved in the training, $l$ denotes the set of locations where the distance between the reconstructed and the true locations is less than 500m. We first analyze the NYCB dataset, as illustrated in Fig.~\ref{Privacy-utility trade-off}(a). As the attack risk increases, the model prediction accuracy also improves. This is because a greater privacy risk corresponds to less noise being added, resulting in a smaller impact on model accuracy. Furthermore, we observe that under equivalent attack risk, the proposed adaptive defense strategy consistently maintains the highest model prediction accuracy. This outcome suggests that our approach can achieve a better trade-off between privacy protection and utility compared to other defense strategies. The strength of the adaptive strategy lies in its ability to flexibly adjust the amount of noise added based on the attack risk, thereby minimizing the impact on model performance while ensuring robust privacy protection. 

Moreover, DPSGD demonstrates significant disadvantages in most cases. Its poor performance can be attributed to the fact that, as a general deep learning privacy protection method, DPSGD lacks specific design considerations for location privacy. In contrast, other methods are more targeted and better equipped to address privacy risks in spatiotemporal federated learning. This highlights the importance of designing dedicated privacy protection mechanisms when handling location data or similar sensitive information to enhance the effectiveness of privacy protection. The analysis results for the Gowalla dataset, presented in Fig.~\ref{Privacy-utility trade-off}(b), further support these conclusions. We observe that, regardless of the dataset, an increase in attack risk correlates with improved model accuracy. At the same time, our adaptive strategy consistently maintains high prediction accuracy across various levels of privacy risk. This indicates that our adaptive defense strategy possesses broad applicability and demonstrates consistently superior performance.

In summary, the consistent results on both datasets show that the proposed defense strategy not only effectively mitigates gradient inversion attacks but also achieves an optimal trade-off between privacy and utility. This finding highlights the effectiveness of our approach in providing robust privacy protection while maximizing model utility, thereby offering valuable theoretical and practical support for future research in privacy protection.

\section{Conclusion and Discussion}
This paper is the first to investigate gradient inversion attacks in spatiotemporal federated learning. We introduce a novel attack method, ST-GIA, specifically designed for this context, which effectively recovers original locations from gradients. Building on this foundation, we further introduce ST-GIA+, which employs an auxiliary model to guide the search for potential locations, enabling more accurate reconstruction of the user's original data. Subsequently, we develop an adaptive differential privacy method to mitigate gradient inversion attacks. Extensive experimental results confirm the effectiveness of the proposed attack methods and defense strategies. In the future, we will explore more spatiotemporal federated learning scenarios, such as traffic flow prediction.

\bibliographystyle{IEEEtran}
\bibliography{refs}

\begin{thebibliography}{10}
\providecommand{\url}[1]{#1}
\csname url@samestyle\endcsname
\providecommand{\newblock}{\relax}
\providecommand{\bibinfo}[2]{#2}
\providecommand{\BIBentrySTDinterwordspacing}{\spaceskip=0pt\relax}
\providecommand{\BIBentryALTinterwordstretchfactor}{4}
\providecommand{\BIBentryALTinterwordspacing}{\spaceskip=\fontdimen2\font plus
\BIBentryALTinterwordstretchfactor\fontdimen3\font minus \fontdimen4\font\relax}
\providecommand{\BIBforeignlanguage}[2]{{%
\expandafter\ifx\csname l@#1\endcsname\relax
\typeout{** WARNING: IEEEtran.bst: No hyphenation pattern has been}%
\typeout{** loaded for the language `#1'. Using the pattern for}%
\typeout{** the default language instead.}%
\else
\language=\csname l@#1\endcsname
\fi
#2}}
\providecommand{\BIBdecl}{\relax}
\BIBdecl

\bibitem{zhang2023federated}
X.~Zhang, Q.~Wang, Z.~Ye, H.~Ying, and D.~Yu, ``Federated representation learning with data heterogeneity for human mobility prediction,'' \emph{IEEE Transactions on Intelligent Transportation Systems}, 2023.

\bibitem{yuan2012discovering}
J.~Yuan, Y.~Zheng, and X.~Xie, ``Discovering regions of different functions in a city using human mobility and pois,'' in \emph{Proceedings of the 18th ACM SIGKDD international conference on Knowledge discovery and data mining}, 2012, pp. 186--194.

\bibitem{liu2017point}
Y.~Liu, C.~Liu, X.~Lu, M.~Teng, H.~Zhu, and H.~Xiong, ``Point-of-interest demand modeling with human mobility patterns,'' in \emph{Proceedings of the 23rd ACM SIGKDD international conference on knowledge discovery and data mining}, 2017, pp. 947--955.

\bibitem{shi2019survey}
Y.~Shi, H.~Feng, X.~Geng, X.~Tang, and Y.~Wang, ``A survey of hybrid deep learning methods for traffic flow prediction,'' in \emph{Proceedings of the 2019 3rd international conference on advances in image processing}, 2019, pp. 133--138.

\bibitem{feng2020pmf}
J.~Feng, C.~Rong, F.~Sun, D.~Guo, and Y.~Li, ``Pmf: A privacy-preserving human mobility prediction framework via federated learning,'' \emph{Proceedings of the ACM on Interactive, Mobile, Wearable and Ubiquitous Technologies}, vol.~4, no.~1, pp. 1--21, 2020.

\bibitem{li2020predicting}
A.~Li, S.~Wang, W.~Li, S.~Liu, and S.~Zhang, ``Predicting human mobility with federated learning,'' in \emph{Proceedings of the 28th International Conference on Advances in Geographic Information Systems}, 2020, pp. 441--444.

\bibitem{fan2024guardian}
M.~Fan, Y.~Liu, C.~Chen, C.~Wang, M.~Qiu, and W.~Zhou, ``Guardian: Guarding against gradient leakage with provable defense for federated learning,'' in \emph{Proceedings of the 17th ACM International Conference on Web Search and Data Mining}, 2024, pp. 190--198.

\bibitem{geiping2020inverting}
J.~Geiping, H.~Bauermeister, H.~Dr{\"o}ge, and M.~Moeller, ``Inverting gradients-how easy is it to break privacy in federated learning?'' \emph{Advances in Neural Information Processing Systems}, vol.~33, pp. 16\,937--16\,947, 2020.

\bibitem{geng2023improved}
J.~Geng, Y.~Mou, Q.~Li, F.~Li, O.~Beyan, S.~Decker, and C.~Rong, ``Improved gradient inversion attacks and defenses in federated learning,'' \emph{IEEE Transactions on Big Data}, 2023.

\bibitem{zhu2019deep}
L.~Zhu, Z.~Liu, and S.~Han, ``Deep leakage from gradients,'' \emph{Advances in neural information processing systems}, vol.~32, 2019.

\bibitem{fan2019decentralized}
Z.~Fan, X.~Song, R.~Jiang, Q.~Chen, and R.~Shibasaki, ``Decentralized attention-based personalized human mobility prediction,'' \emph{Proceedings of the ACM on Interactive, Mobile, Wearable and Ubiquitous Technologies}, vol.~3, no.~4, pp. 1--26, 2019.

\bibitem{wang2022location}
S.~Wang, B.~Wang, S.~Yao, J.~Qu, and Y.~Pan, ``Location prediction with personalized federated learning,'' \emph{Soft Computing}, pp. 1--12, 2022.

\bibitem{zhao2020idlg}
B.~Zhao, K.~R. Mopuri, and H.~Bilen, ``idlg: Improved deep leakage from gradients,'' \emph{arXiv preprint arXiv:2001.02610}, 2020.

\bibitem{wang2020sapag}
Y.~Wang, J.~Deng, D.~Guo, C.~Wang, X.~Meng, H.~Liu, C.~Ding, and S.~Rajasekaran, ``Sapag: A self-adaptive privacy attack from gradients,'' \emph{arXiv preprint arXiv:2009.06228}, 2020.

\bibitem{wei2020framework}
W.~Wei, L.~Liu, M.~Loper, K.-H. Chow, M.~E. Gursoy, S.~Truex, and Y.~Wu, ``A framework for evaluating gradient leakage attacks in federated learning,'' \emph{arXiv preprint arXiv:2004.10397}, 2020.

\bibitem{yin2021see}
H.~Yin, A.~Mallya, A.~Vahdat, J.~M. Alvarez, J.~Kautz, and P.~Molchanov, ``See through gradients: Image batch recovery via gradinversion,'' in \emph{Proceedings of the IEEE/CVF conference on computer vision and pattern recognition}, 2021, pp. 16\,337--16\,346.

\bibitem{xue2022translating}
H.~Xue, F.~D. Salim, Y.~Ren, and C.~L. Clarke, ``Translating human mobility forecasting through natural language generation,'' in \emph{Proceedings of the Fifteenth ACM International Conference on Web Search and Data Mining}, 2022, pp. 1224--1233.

\bibitem{xue2022leveraging}
H.~Xue, B.~P. Voutharoja, and F.~D. Salim, ``Leveraging language foundation models for human mobility forecasting,'' in \emph{Proceedings of the 30th International Conference on Advances in Geographic Information Systems}, 2022, pp. 1--9.

\bibitem{chang2023llm4ts}
C.~Chang, W.-C. Peng, and T.-F. Chen, ``Llm4ts: Two-stage fine-tuning for time-series forecasting with pre-trained llms,'' \emph{arXiv preprint arXiv:2308.08469}, 2023.

\bibitem{li2024large}
P.~Li, M.~de~Rijke, H.~Xue, S.~Ao, Y.~Song, and F.~D. Salim, ``Large language models for next point-of-interest recommendation,'' \emph{arXiv preprint arXiv:2404.17591}, 2024.

\bibitem{wang2023would}
X.~Wang, M.~Fang, Z.~Zeng, and T.~Cheng, ``Where would i go next? large language models as human mobility predictors,'' \emph{arXiv preprint arXiv:2308.15197}, 2023.

\bibitem{hochreiter1997long}
S.~Hochreiter and J.~Schmidhuber, ``Long short-term memory,'' \emph{Neural computation}, vol.~9, no.~8, pp. 1735--1780, 1997.

\bibitem{feng2018deepmove}
J.~Feng, Y.~Li, C.~Zhang, F.~Sun, F.~Meng, A.~Guo, and D.~Jin, ``Deepmove: Predicting human mobility with attentional recurrent networks,'' in \emph{Proceedings of the 2018 world wide web conference}, 2018, pp. 1459--1468.

\bibitem{liu2016predicting}
Q.~Liu, S.~Wu, L.~Wang, and T.~Tan, ``Predicting the next location: A recurrent model with spatial and temporal contexts,'' in \emph{Proceedings of the AAAI conference on artificial intelligence}, vol.~30, no.~1, 2016.

\bibitem{bassily2015local}
R.~Bassily and A.~Smith, ``Local, private, efficient protocols for succinct histograms,'' in \emph{Proceedings of the forty-seventh annual ACM symposium on Theory of computing}, 2015, pp. 127--135.

\bibitem{wang2017locally}
T.~Wang, J.~Blocki, N.~Li, and S.~Jha, ``Locally differentially private protocols for frequency estimation,'' in \emph{26th USENIX Security Symposium (USENIX Security 17)}, 2017, pp. 729--745.

\bibitem{dwork2006differential}
C.~Dwork, ``Differential privacy,'' in \emph{International colloquium on automata, languages, and programming}.\hskip 1em plus 0.5em minus 0.4em\relax Springer, 2006, pp. 1--12.

\bibitem{4389483}
F.~McSherry and K.~Talwar, ``Mechanism design via differential privacy,'' in \emph{48th Annual IEEE Symposium on Foundations of Computer Science (FOCS'07)}, 2007, pp. 94--103.

\bibitem{cao2020pglp}
Y.~Cao, Y.~Xiao, S.~Takagi, L.~Xiong, M.~Yoshikawa, Y.~Shen, J.~Liu, H.~Jin, and X.~Xu, ``Pglp: Customizable and rigorous location privacy through policy graph,'' in \emph{European Symposium on Research in Computer Security}.\hskip 1em plus 0.5em minus 0.4em\relax Springer, 2020, pp. 655--676.

\bibitem{luo2021stan}
Y.~Luo, Q.~Liu, and Z.~Liu, ``Stan: Spatio-temporal attention network for next location recommendation,'' in \emph{Proceedings of the web conference 2021}, 2021, pp. 2177--2185.

\bibitem{abadi2016deep}
M.~Abadi, A.~Chu, I.~Goodfellow, H.~B. McMahan, I.~Mironov, K.~Talwar, and L.~Zhang, ``Deep learning with differential privacy,'' in \emph{Proceedings of the 2016 ACM SIGSAC conference on computer and communications security}, 2016, pp. 308--318.

\bibitem{andres2013geo}
M.~E. Andr{\'e}s, N.~E. Bordenabe, K.~Chatzikokolakis, and C.~Palamidessi, ``Geo-indistinguishability: Differential privacy for location-based systems,'' in \emph{Proceedings of the 2013 ACM SIGSAC conference on Computer \& communications security}, 2013, pp. 901--914.

\bibitem{takagi2020geo}
S.~Takagi, Y.~Cao, Y.~Asano, and M.~Yoshikawa, ``Geo-graph-indistinguishability: Location privacy on road networks based on differential privacy,'' \emph{arXiv preprint arXiv:2010.13449}, 2020.

\end{thebibliography}

\begin{IEEEbiography}[{\includegraphics[width=1in,height=1.25in, clip, keepaspectratio]{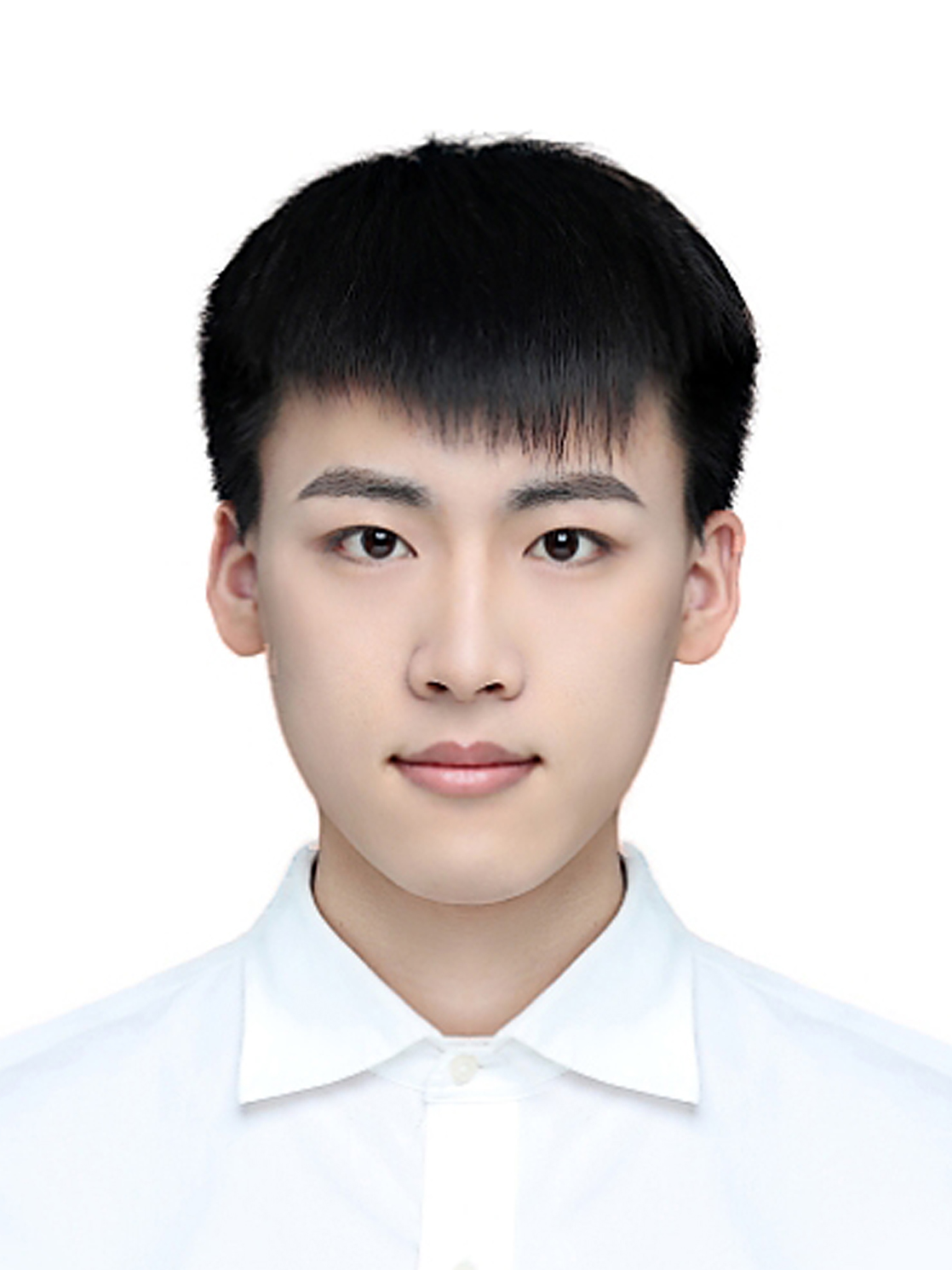}}]{Lele Zheng} received the B.S. degree from Xidian University, China, in 2018, where he is currently pursuing the Ph.D. degree with the School of Computer Science and Technology. His research interests include differential privacy and federated learning.
\end{IEEEbiography}

\begin{IEEEbiography}[{\includegraphics[width=1in,height=1.25in,clip,keepaspectratio]{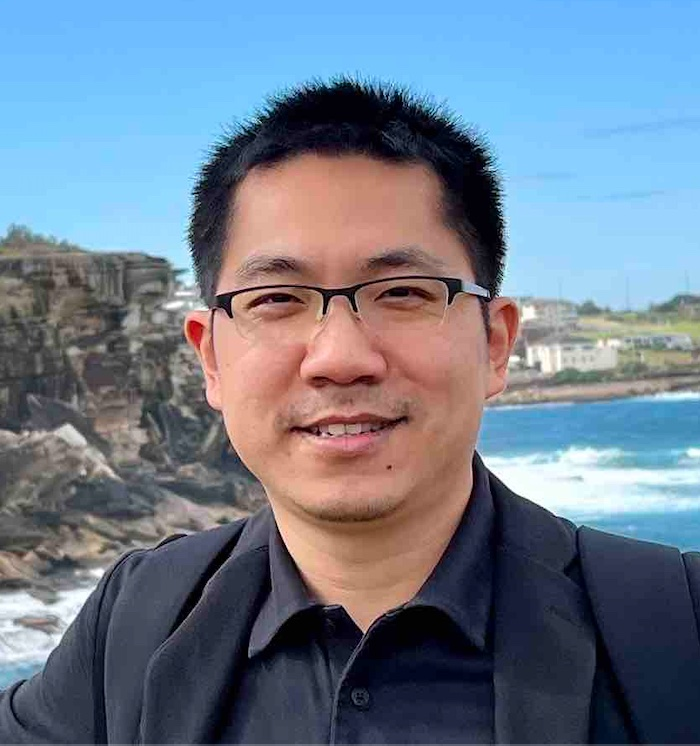}}]{Yang Cao}
is an Associate Professor at the Department of Computer Science, Tokyo Institute of Technology (Tokyo Tech), and directing the Trustworthy Data Science Lab. He is passionate about studying and teaching on algorithmic trustworthiness in data science and AI. Two of his papers on data privacy were selected as best paper finalists in top-tier conferences IEEE ICDE 2017 and ICME 2020. He was a recipient of the IEEE Computer Society Japan Chapter Young Author Award 2019, Database Society of Japan Kambayashi Young Researcher Award 2021. His research projects were/are supported by JSPS, JST, MSRA, KDDI, LINE, WeBank, etc.
\end{IEEEbiography}

\begin{IEEEbiography}[{\includegraphics[width=1in,height=1.25in,clip,keepaspectratio]{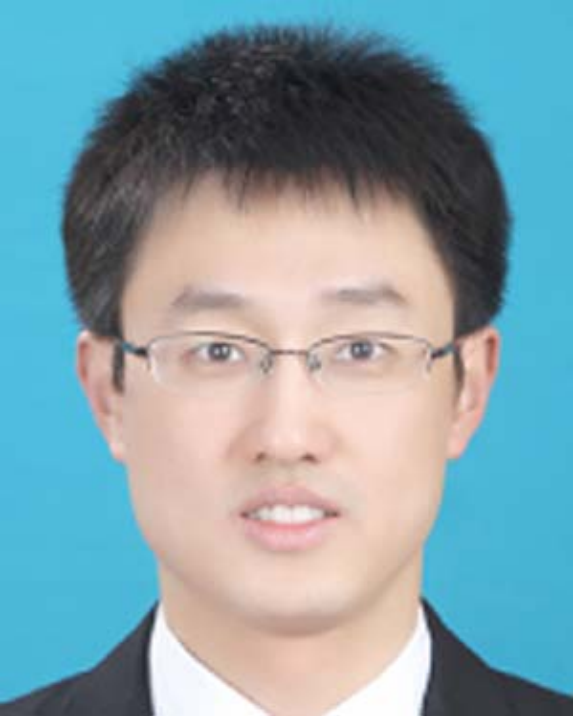}}]{Renhe Jiang} (Member, IEEE) received the BS degree in software engineering from the Dalian University of Technology, China, in 2012, the MS degree in information science from Nagoya University, Japan, in 2015, and the PhD degree in civil engineering from The University of Tokyo, Japan, in 2019. From 2019, he became an assistant professor with Information Technology Center, The University of Tokyo. His research interests include ubiquitous computing, deep learning, and spatio-temporal data analysis.
\end{IEEEbiography}

\begin{IEEEbiography}[{\includegraphics[width=1in,height=1.25in,clip,keepaspectratio]{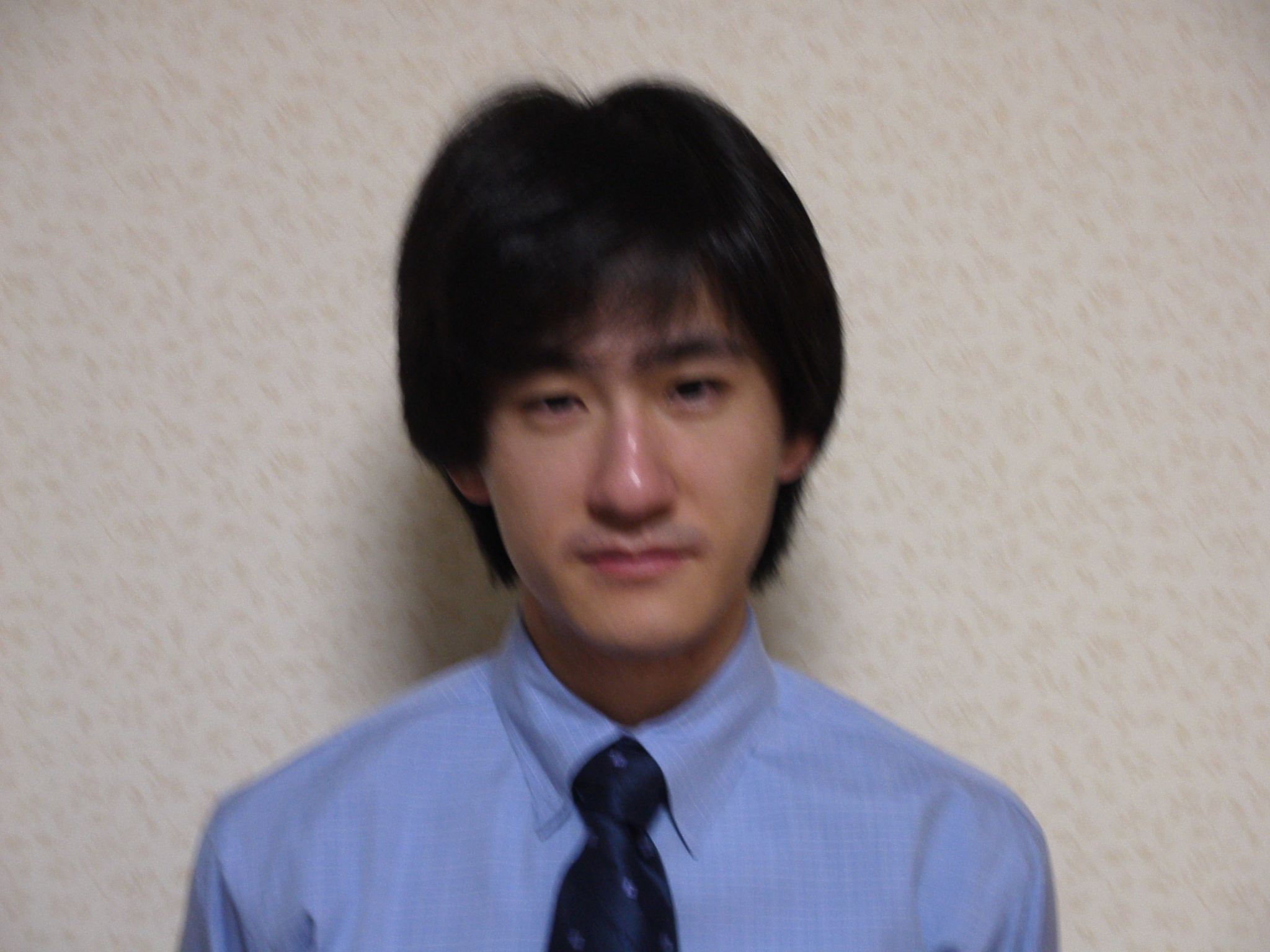}}]{Kenjiro Taura} is associate professor at Department of Information and Communication Engineering, University of Tokyo. He was born in 1969, and received his B.S., M.S., and DSc degrees from University of Tokyo in 1992, 1994, and 1997. His major research interests are centered around parallel/distributed computing and programming languages. His expertise includes efficient dynamic load balancing, parallel and distributed garbage collection, and parallel/distributed workflow systems. He is a member of ACM and IEEE.
\end{IEEEbiography}

\begin{IEEEbiography}[{\includegraphics[width=1in,height=1.25in,clip,keepaspectratio]{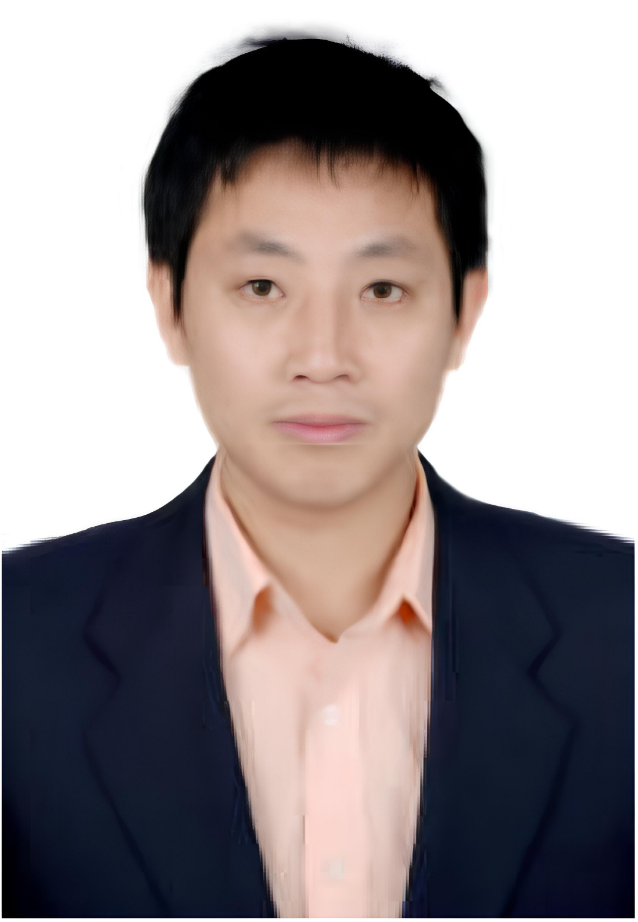}}]{Yulong Shen}  (Member, IEEE) received the B.S. and M.S. degrees in computer science and the Ph.D. degree in cryptography from Xidian University, Xi'an, China, in 2002, 2005, and 2008, respectively. He is currently a Professor with the School of Computer Science and Technology, Xidian University, where he is also an Associate Director of the Shaanxi Key Laboratory of Network and System Security and a member of the State Key Laboratory of Integrated Services Networks. His research interests include wireless network security and cloud computing security. He has also served on the technical program committees of several international conferences, including ICEBE, INCoS, CIS, and SOWN.
\end{IEEEbiography}

\begin{IEEEbiography}[{\includegraphics[width=1in,height=1.25in,clip,keepaspectratio]{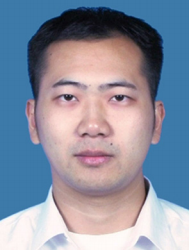}}]{Sheng Li} received his BS and ME degrees in 2006 and 2009 from Nanjing University, Nanjing, China, and his Ph.D. from Kyoto University, Kyoto, Japan, in 2016. From 2009 to 2012, he worked at the joint lab of the Chinese University Hong Kong and Shenzhen City, researching speech technology-assisted language learning. From 2016 to 2017, he worked as a researcher at Kyoto University, studying speech recognition systems for humanoid robots. In 2017, he joined the National Institute of Information and Communications Technology, Kyoto, Japan, as a researcher working on speech recognition. He served as workshop co-organizer and chair in interspeech2020, coling2022, odyssey2022, ACM Multimedia Asia2023/2024, and ICASSP2024. He is a member of the Acoustic Society of Japan (ASJ), the International Speech Communication Association (ISCA), and IEEE. He is now a member of the Speech, Language, and Audio (SLA) Technical Committee for APSIPA.

\end{IEEEbiography}

\begin{IEEEbiography}[{\includegraphics[width=1in,height=1.25in,clip,keepaspectratio]{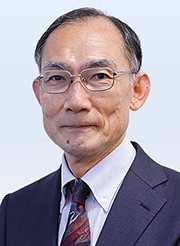}}]{Masatoshi Yoshikawa} is a Professor at Osaka Seikei University. He received the B.E., M.E. and Dr. Eng. degrees from Department of Information Science, Kyoto University in 1980, 1982 and 1985, respectively. He is a Professor Emeritus at Kyoto University. His current research interests include databases and privacy-enhancing technologies. He is a members of the IEEE ICDE Steering Committee.
\end{IEEEbiography}

\end{document}